\documentclass[acmsmall]{acmart}
\AtBeginDocument{%
  }

\setcopyright{acmlicensed}
\copyrightyear{2025}
\acmYear{2025}
\acmDOI{XXXXXXX.XXXXXXX}
\acmISBN{978-1-4503-XXXX-X/2025/12}

\usepackage{graphicx}%
\usepackage{multirow}%
\usepackage{amsmath,amsfonts}%
\usepackage{amsthm}%
\usepackage{mathrsfs}%
\usepackage[title]{appendix}%
\usepackage{xcolor}%
\usepackage{textcomp}%
\usepackage{manyfoot}%
\usepackage{booktabs}%
\usepackage{algorithm}%
\usepackage{algorithmicx}%
\usepackage{algpseudocode}%
\usepackage{listings}%

\usepackage{framed}
\usepackage{makecell}
\usepackage{multirow}

\usepackage{subcaption}
\usepackage{lmodern}

\definecolor{colYes}{HTML}{1A854D}     
\definecolor{colNo}{HTML}{D13B40}      
\definecolor{colPartial}{HTML}{E89B36} 

\newcommand{\yes}{\textcolor{colYes}{\textbf{Yes}}}
\newcommand{\no}{\textcolor{colNo}{\textbf{No}}}
\newcommand{\partialc}{\textcolor{colPartial}{\textbf{Partial}}} 

\begin{document}

\title{A Systematic Survey on Large Language Models for Algorithm Design}






\author{Fei Liu}
\email{fliu36-c@my.cityu.edu.hk}
\affiliation{%
  \institution{City University of Hong Kong}
  \city{Hong Kong}
  \country{China}
}

\author{Yiming Yao}
\email{yimingyao3-c@my.cityu.edu.hk}
\affiliation{%
  \institution{City University of Hong Kong}
  \city{Hong Kong}
  \country{China}
}

\author{Ping Guo}
\email{pingguo5-c@my.cityu.edu.hk}
\affiliation{%
  \institution{City University of Hong Kong}
  \city{Hong Kong}
  \country{China}
}

\author{Zhiyuan Yang}
\email{zhiyuan.yang@my.cityu.edu.hk}
\affiliation{%
  \institution{Huawei Noah’s Ark Lab}
  \city{Hong Kong}
  \country{China}
}

\author{Xi Lin}
\email{xi.lin@my.cityu.edu.hk}
\affiliation{%
  \institution{Xi'an Jiaotong University}
  \city{Xi'an}
  \country{China}
}

\author{Zhe Zhao}
\email{zzhao26-c@my.cityu.edu.hk}
\affiliation{%
  \institution{City University of Hong Kong}
  \city{Hong Kong}
  \country{China}
}

\author{Xialiang Tong}
\email{tongxialiang@huawei.com}
\affiliation{%
  \institution{Huawei Noah’s Ark Lab}
  \city{Shenzhen}
  \country{China}
}

\author{Kun Mao}
\email{maokun@huawei.com}
\affiliation{%
  \institution{Huawei Cloud EI Service Product Dept.}
  \city{Shenzhen}
  \country{China}
}

\author{Zhichao Lu}
\email{luzhichaocn@gmail.com}
\affiliation{%
  \institution{City University of Hong Kong}
  \city{Hong Kong}
  \country{China}
}

\author{Zhenkun Wang}
\email{wangzk3@sustech.edu.cn}
\affiliation{%
  \institution{Southern University of Science and Technology}
  \city{Shenzhen}
  \country{China}
}

\author{Mingxuan Yuan}
\email{yuan.mingxuan@huawei.com}
\affiliation{%
  \institution{Huawei Noah’s Ark Lab}
  \city{Hong Kong}
  \country{China}
}

\author{Qingfu Zhang}
\authornote{the corresponding author}
\email{qingfu.zhang@cityu.edu.hk}
\affiliation{%
  \institution{City University of Hong Kong}
  \city{Hong Kong}
  \country{China}
}

\settopmatter{printacmref=false}
\setcopyright{none}
\renewcommand{\shortauthors}{Liu et al.}



\begin{abstract}
Algorithm design is crucial for effective problem-solving across various domains. The advent of Large Language Models (LLMs) has notably enhanced the automation and innovation within this field, offering new perspectives and promising solutions. In just a few years, this integration has yielded remarkable progress in areas ranging from combinatorial optimization to scientific discovery. Despite this rapid expansion, a holistic understanding of the field is hindered by the lack of a systematic review, as existing surveys either remain limited to narrow sub-fields or with different objectives. This paper seeks to provide a systematic review of algorithm design with LLMs. We introduce a taxonomy that categorises the roles of LLMs as optimizers, predictors, extractors and designers, analyzing the progress, advantages, and limitations within each category. We further synthesize literature across the three phases of the algorithm design pipeline and across diverse algorithmic applications that define the current landscape. Finally, we outline key open challenges and opportunities to guide future research. To support future research and collaboration, we provide an accompanying repository at: \url{https://github.com/FeiLiu36/LLM4AlgorithmDesign}.
\end{abstract}




\received{20 February 2007}
\received[revised]{12 March 2009}
\received[accepted]{5 June 2009}

\maketitle

\section{Introduction}

Algorithms, which are a sequence of computational steps for solving a well-specified computational problem~\citep{cormen2022introduction}, play a crucial role in addressing various problems across various domains such as industry, economics, healthcare, and technology~\citep{kleinberg2006algorithm,cormen2022introduction}. Traditionally, designing algorithm has been a labor-intensive process that demands deep expertise. Recently, there has been a surge in interest towards employing learning and computational intelligence methods techniques to enhance and automate the algorithm development process~\citep{bengio2021machine,tang2024learn}.

Over the past few years, the application of Large Language Models for Algorithm Design (LLM4AD) has merged as a promising research area with the potential to fundamentally transform the ways in which algorithms are designed, implemented, and optimized. The remarkable capability and flexibility of LLMs have demonstrated potential in enhancing the algorithm design process from algorithm ideation~\citep{girotra2023ideas} to implementation~\citep{liu2024evolution}. This approach not only reduces the human effort required in the design phase but also enhances the creativity and efficiency of the produced solutions~\citep{romera2024mathematical,liu2024llm4ad,novikov2025alphaevolve}.

Despite this surge of interest, the field lacks a systematic survey with a clear organizational structure. While several recent surveys have reviewed the interaction of LLM and algorithms, they either focus on narrow sub-fields or adjacent but different domains. For example, code generation surveys~\citep{wang2023review,zheng2023survey,jiang2024survey,wang2024survey} emphasize translating specifications into executable code, focusing on implementation rather than the upstream ideation and strategic design of algorithms. Reviews on optimization and evolutionary computation~\citep{wu2024evolutionary,fan2024artificial,zhang2025systematic,huang2024large} target specific problem classes (e.g., combinatorial optimization), offering limited breadth beyond the target sub-fields. Surveys on LLM agents and planning~\citep{pallagani2024prospects,wang2024survey,guo2024large,ferrag2025llm} center on system architecture, tool use, and reasoning pipelines rather than the creation and refinement of core algorithmic logic. As summarized in Table \ref{tab:llm_survey_colored}, these surveys cover partial stages of algorithm design or specific algorithm types.

This paper aims to fill this gap by providing a systematic survey dedicated to the emerging field of LLM4AD. To ensure conceptual clarity, we first establish a precise definition for ``algorithm design'' and the scope of this survey, distinguishing it from general-purpose programming. We then introduce a role-based taxonomy that organizes the literature according to the four fundamental ways LLMs are being employed in the algorithm design process. By synthesizing insights from over 180 recent papers, we analyze the using of LLM in different algorithm design stages and application domains. Finally, we critically assess the field's current challenges and identify promising opportunities for future research. We intend for this survey to serve as an essential resource for both newcomers seeking a structured overview and experts looking for a consolidated analysis of the latest advancements.

\begin{table*}[t]
\centering
\caption{A Comparison of Different Survey Papers Across Algorithmic Design Stages and Algorithm Types.}
\label{tab:llm_survey_colored}
\resizebox{0.88\textwidth}{!}{%
\begin{tabular}{@{}lcccccccc@{}}
\toprule
\multirow{3}{*}{\textbf{Year}} & 
\multirow{3}{*}{\textbf{Survey}} & 
\multirow{3}{*}{\textbf{LLM}} & \multicolumn{3}{c}{\textbf{Algorithmic Design Stages}} & \multicolumn{3}{c}{\textbf{Algorithm Types}} \\
\cmidrule(lr){4-6} \cmidrule(l){7-9} 
& & & \begin{tabular}[c]{@{}c@{}}Algorithm \\ Ideation\end{tabular} & \begin{tabular}[c]{@{}c@{}}Algorithm \\ Implementation\end{tabular} & \begin{tabular}[c]{@{}c@{}}Algorithm \\ Evaluation\end{tabular} & Non-heuristic & \begin{tabular}[c]{@{}c@{}}Heuristic and \\ Metaheuristic\end{tabular} & \begin{tabular}[c]{@{}c@{}}Hybrid\end{tabular} \\
\midrule
\multirow{3}{*}{2023} 
& Wang et al. \citep{wang2023review}   & \yes      & \no       & \yes      & \partialc & \no       & \no       & \no       \\
& Zhao et al. \citep{zhao2023automated}   & \yes      & \yes      & \yes      & \partialc & \no       & \yes      & \partialc \\
& Zheng et al. \citep{zheng2023survey}  & \yes      & \no       & \yes      & \partialc & \partialc & \partialc & \partialc \\
\midrule
\multirow{6}{*}{2024} 
& Ahn et al. \citep{ahn2024large}    & \yes      & \partialc & \no       & \no       & \yes      & \no       & \no       \\
& Guo et al. \citep{guo2024large}    & \yes      & \no       & \partialc & \partialc & \partialc & \partialc & \partialc \\
& Jiang et al. \citep{jiang2024survey}  & \yes      & \no       & \yes      & \partialc & \no       & \no       & \no       \\
& Joel et al. \citep{joel2024survey}   & \yes      & \no       & \yes      & \partialc & \no       & \no       & \no       \\
& Wu et al. \citep{wu2024evolutionary}     & \yes      & \partialc & \partialc & \partialc & \no       & \partialc & \partialc \\
& Fan et al. \citep{fan2024artificial}    & \partialc & \no       & \yes      & \yes      & \yes      & \yes      & \yes      \\
\midrule
\multirow{5}{*}{2025} 
& Ferrag et al. \citep{ferrag2025llm} & \yes      & \no       & \partialc & \partialc & \partialc & \partialc & \partialc \\
& Zhang et al. \citep{zhang2025systematic} & \yes      & \partialc & \partialc & \partialc & \no       & \partialc & \partialc \\
& Ma et al. \citep{ma2025toward}    & \partialc & \partialc & \yes      & \yes      & \no       & \partialc & \partialc \\
& Da Ros et al. \citep{da2025large} & \yes      & \partialc & \yes      & \yes      & \no       & \partialc & \partialc \\
\midrule
\multicolumn{2}{l}{\textbf{Ours}} & \yes & \yes & \yes & \yes & \yes & \yes & \yes \\
\bottomrule
\end{tabular}%
} 
\end{table*}

\begin{figure}[h]
    \centering    
    \includegraphics[width=0.83\linewidth]{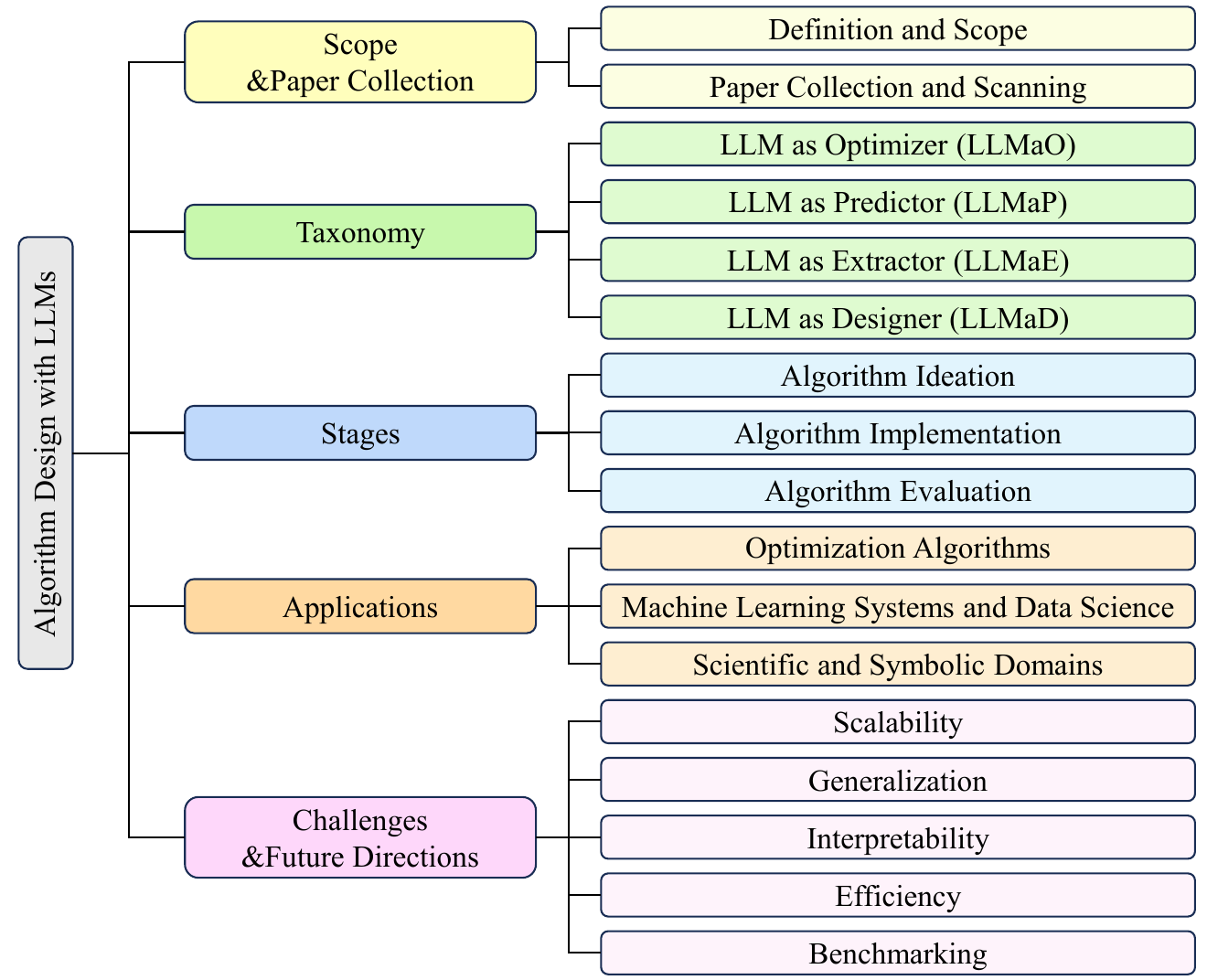}
    \Description{}
    \caption{Overview of the Survey Structure: Scope, Taxonomy of LLM Roles, Stages of Algorithm Design, Applications, Open Challenges and Future Directions.}
    \label{fig:structure}
\end{figure}

Fig.~\ref{fig:structure} provides an overview of this survey's structure. The remainder of the paper is organized as follows. Section~\ref{sec2:method} outlines our survey methodology, including scope and literature collection and screening pipeline. Section~\ref{sec3:taxonomy} introduces a taxonomy for organizing LLM4AD research, categorizing works based on the primary role of the LLM in the algorithm design process: as optimizer (LLMaO), predictor (LLMaP), extractor (LLMaE), or designer (LLMaD). Subsequently, Section~\ref{sec4:stage} reviews works across different algorithm development stages, while Section~\ref{sec5:appli} summarizes key application domains. Section~\ref{sec6:challenges} discusses current open challenges and promising future research directions. Finally, Section~\ref{sec7:conclusion} presents the conclusions.

\section{Methodology}\label{sec2:method}
\subsection{Definition and Scope}

This paper focuses on studies where LLMs substantively contribute to the conception, synthesis, or refinement of algorithms. This section defines our core concepts and delineates the scope of this survey to clarify its boundaries with respect to related research works.

\paragraph{Algorithm}
We adopt the definition of an ``algorithm'' as a well-defined computational procedure that transforms a set of inputs into a set of outputs in a finite amount of time \citep{cormen2022introduction}. This includes deterministic and stochastic procedures, exact methods and approximations~\citep{kleinberg2006algorithm}, as well as heuristics and metaheuristics~\citep{glover2006handbook}. Examples include a procedure that sorts a set of integers, a method that finds a shortest path on a graph, or a heuristic for a scheduling problem. This scope covers both classic textbook algorithms and practically motivated strategies that trade optimality for speed or simplicity.

\paragraph{Algorithm Design}
We cover three stages of algorithm design: i) generation of algorithmic ideas or pseudocodes~\citep{si2024can}; ii) algorithm implementation, which concerns the production of executable code from specifications~\citep{van2025code}; and iii) algorithm evaluation, where LLMs are used for assessing the performance and analyzing the behavior of the designed algorithms~\citep{liu2024llm4ad}. A crucial distinction is made between algorithm design and general-purpose code generation~\citep{jiang2024survey}. Studies that merely translate algorithms or procedures into code are excluded.

\paragraph{Large Language Models}
Our focus is on large-scale language models, typically with billions of parameters, capable of sophisticated text and code processing~\citep{zhao2023survey}. This includes both text-only and multi-modal LLMs where language is a core component. We exclude smaller-scale models and traditional machine learning approaches for algorithm generation, which have been discussed by other survey papers~\cite{bengio2021machine,ma2025toward}.

\subsection{Paper Collection and Scanning}

We introduce the detailed pipeline for paper collection and scanning, which consists of three stages:

\begin{itemize}

\item \textbf{Stage I Data Extraction and Collection:} We collect the related papers through Google Scholar, Web of Science, and Scopus. The logic of our search is the title must include any combinations of at least one of the following two groups of words {``LLM'', ``LLMs'', ``Large Language Model'', ``Large Language Models''} and {``Algorithm'', ``Heuristic'',  ``Search'', ``Optimization'', ``Optimizer'', ``Design'', ``Function''} (e.g., LLM and optimization, LLMs and algorithm). After removing duplicate papers, we ended up with around 3,000 papers as of October 1, 2025.

\begin{figure}[t]
\centering
\includegraphics[width=0.98\linewidth]{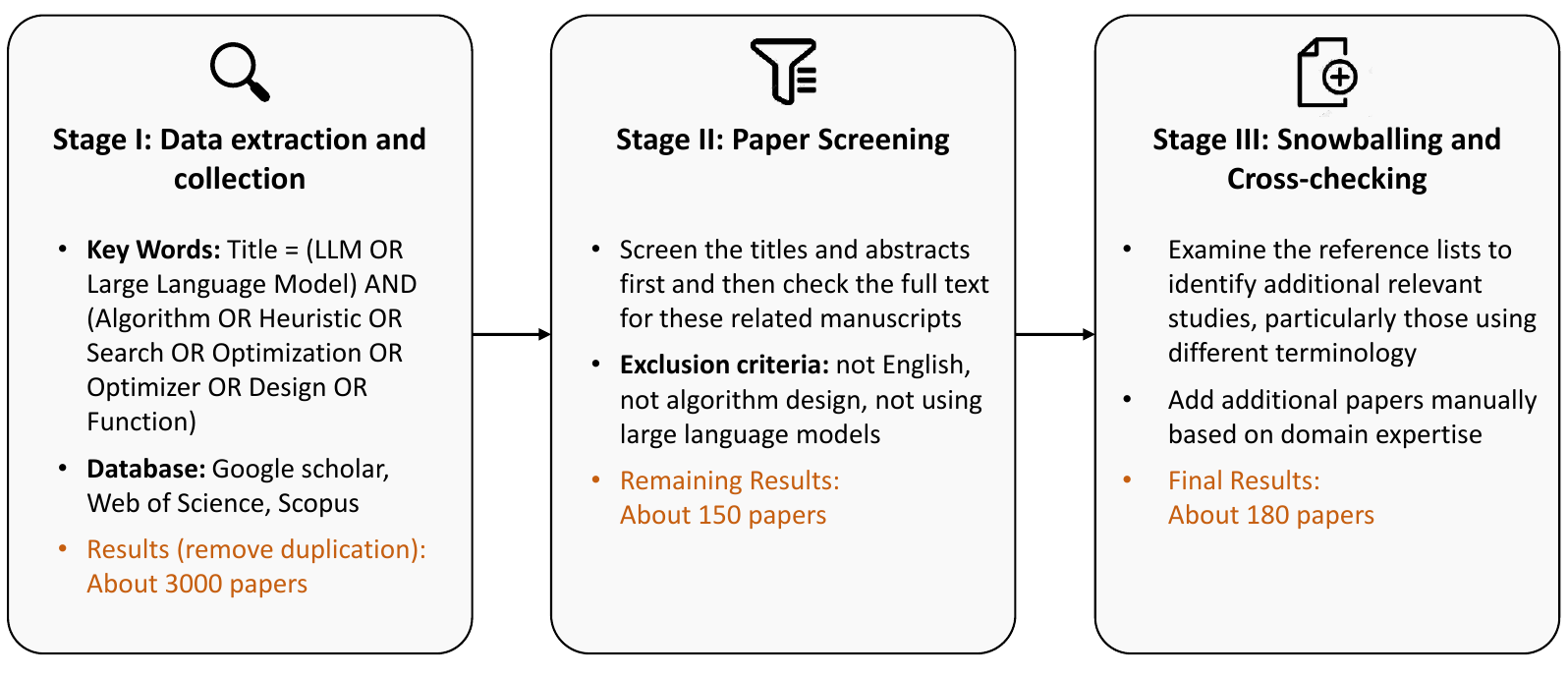}
\Description{Four stages for paper collection.}
\caption{A Three-stage Pipeline for Paper Collection and Screening.}
\label{fig:methodology}
\end{figure}

\begin{figure}[htbp]
\centering
\begin{subfigure}[t]{0.4\textwidth}
\centering
\includegraphics[width=\textwidth]{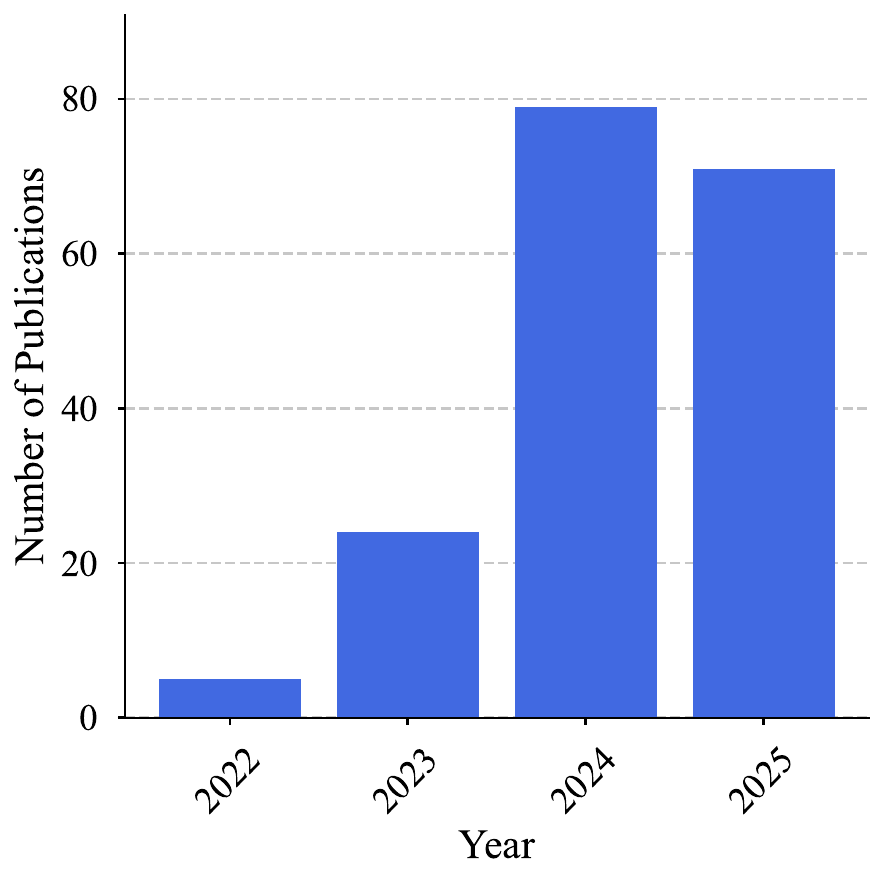}
\caption{Annual Publication Volume of Surveyed Papers (until Sep. 2025).}
\label{fig:number_of_publications}
\end{subfigure}
\hfill 
\begin{subfigure}[t]{0.58\textwidth}
\centering
\includegraphics[width=\textwidth]{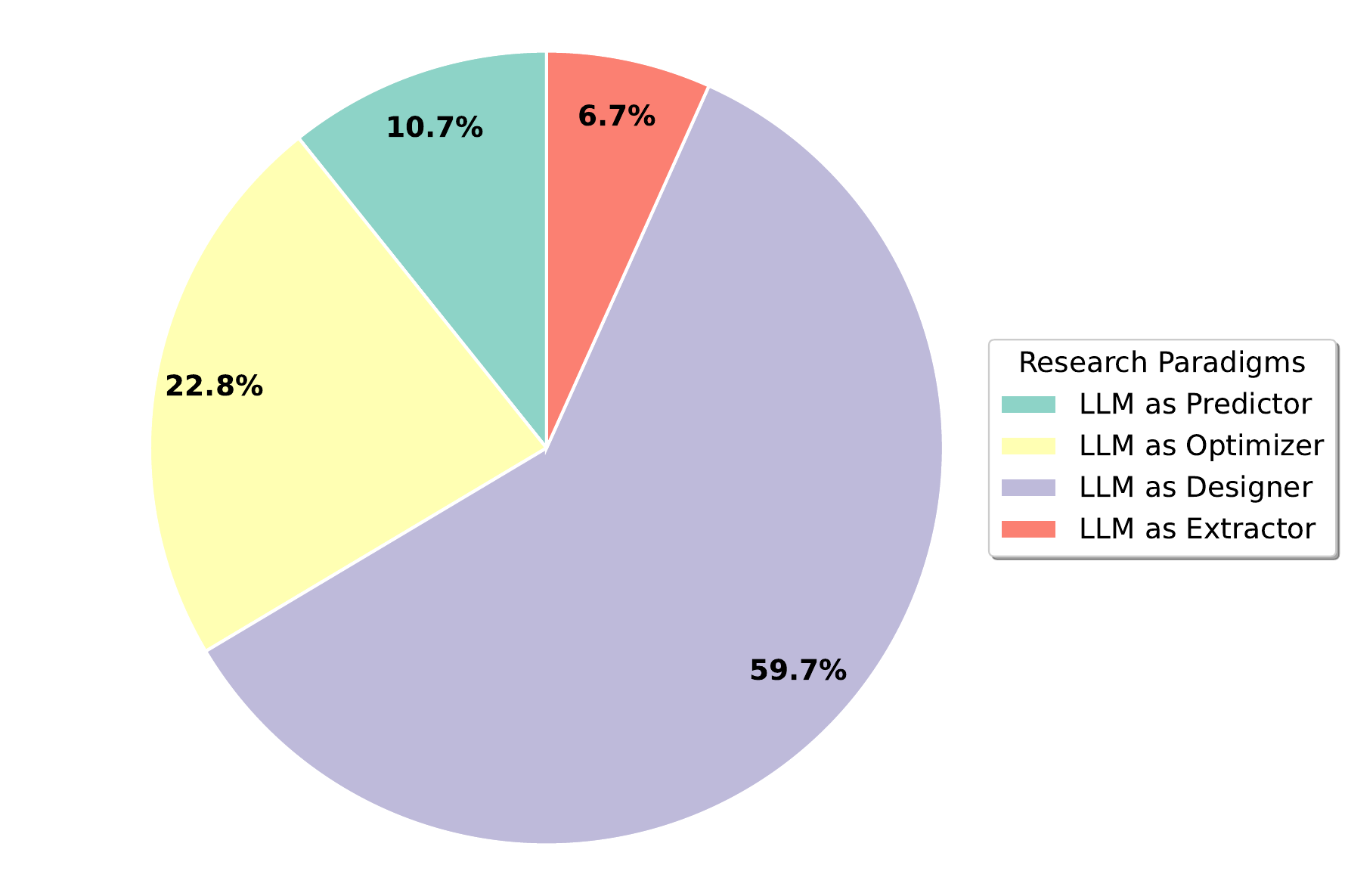}
\caption{Distribution on Research Paradigms}
\label{fig:paradigm_distribution}
\end{subfigure}
\caption{Publication Trends and Paradigm Distribution in the Surveyed Literature.}
\label{fig:publication_overview}
\end{figure}

\item \textbf{Stage II: Paper Screening}
This stage involved a two-step screening process to identify the most relevant papers. First, we screened the titles and abstracts of the 3,000 papers against predefined \textbf{exclusion criteria}: i) The paper is \textit{not} written in English. ii) The paper's primary focus is \textit{not} on algorithm design (e.g., it focuses only on general code generation without any algorithmic component). iii) The paper does \textit{not} utilize large language models as defined in our scope. This initial screening narrowed the corpus down to about 500 papers. Subsequently, we conducted a full-text review of these manuscripts, applying the same exclusion criteria more rigorously to filter out papers that, upon closer inspection, lacked substantive content on LLM-aided algorithm design. This thorough review resulted in a refined set of 150 high-quality, relevant papers.

\item \textbf{Stage III: Snowballing and Cross-checking}
To ensure comprehensive coverage and mitigate the limitations of keyword-based searches, we performed a backward snowballing procedure on the 150 papers. This involved manually examining the reference lists of these papers to identify relevant studies that our initial search may have missed (for instance, papers using terminology like ``code generation'' instead of ``algorithm design'' but still involving some algorithm design tasks). Additionally, we manually appended a small number of works based on the authors' domain knowledge to avoid omitting any important contributions. After integrating these additional papers identified through snowballing and expert knowledge, our final corpus contained over 180 papers.
\end{itemize}

We acknowledge that, given the vast domains of algorithm design and the volume of literature, it is impossible to guarantee an exhaustive coverage of all relevant papers. Instead, our objective is to systematically survey the landscape, focusing on a representative body of work that allows us to organize and discuss the field. Fig.~\ref{fig:number_of_publications} illustrates the number of publications per year surveyed in this paper. The graph shows a marked rise in research activity related to LLM4AD, and most of the related studies have been conducted in the last two years.


\section{Taxonomy of LLM Roles in Algorithm Design}\label{sec3:taxonomy}

According to the roles of LLM in algorithm design, existing works can be categorized into four paradigms: LLM as Optimizer (LLMaO), LLM as Predictor (LLMaP), LLM as Extractor (LLMaE), and LLM as Designer (LLMaD). Fig.~\ref{fig:paradigm_distribution} displays the distribution of publications across these four paradigms. This section discusses the progress, advantages, and limitations of each category.

\subsection{LLM as Optimizer (LLMaO)}

\begin{figure}[t]
    \centering
    \includegraphics[width=0.9\linewidth]
    {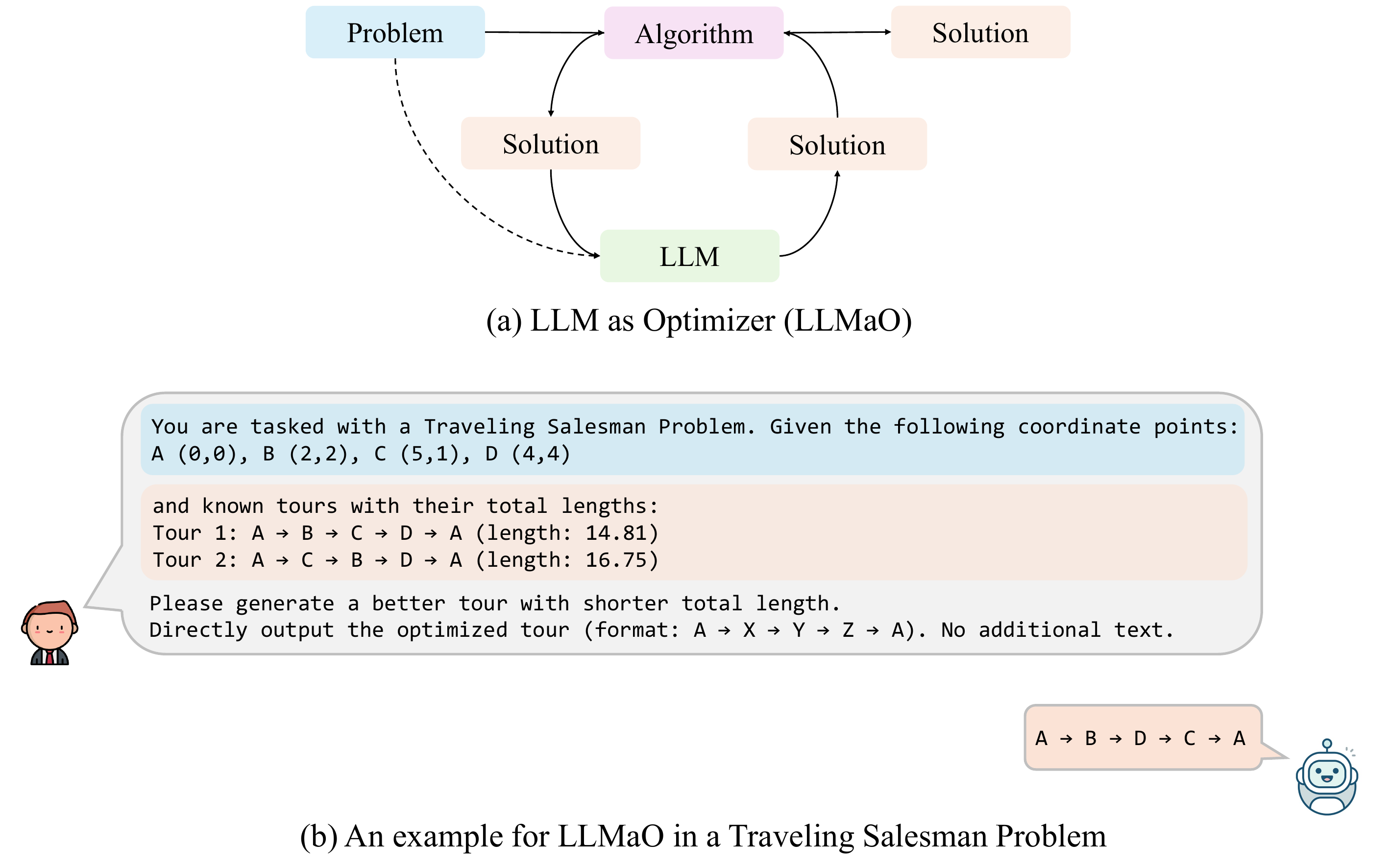}
    \Description{}
    \caption{(a) Large Language Models as Optimizers (LLMaO). LLMs Serve as Optimizers within the Algorithm to Generate New Solutions. (b) An Example for LLMaO on Traveling Salesman Problem.}
    \label{fig:llmao}
\end{figure}

In LLMaO (Fig.~\ref{fig:llmao}), LLMs are employed as a black-box optimizer within an algorithmic framework to generate and refine solutions. Fig.~\ref{fig:llmao} (b) illustrates an example applied to the Traveling Salesman Problem (TSP), which involves optimizing a tour that visits each city (node) exactly once and returns to the starting city, with the goal of minimizing the total route length. In this example, we are given four nodes (A, B, C, and D) with their coordinates, and the algorithm has two existing tours (tour 1 and tour 2) each with a different sequence and route length. The LLM is used to generate a potentially better tour with a shorter length. The LLM directly outputs the optimized tour sequence, which is commonly the role of hard-coded optimizers within traditional algorithmic approaches.


One of the initial efforts to utilize LLM as Optimizer in algorithm design is by \citet{yang2024large}. They leverage the in-context learning capabilities of LLMs to generate new solutions for specific problems based on previously evaluated solutions. This method is applied iteratively to refine solutions further. \citet{yang2024large} have successfully demonstrated this technique across various domains, including continuous and combinatorial optimization, as well as machine learning tasks.

From an evolutionary algorithm perspective, using LLMs to generate solutions from existing data can be seen as analogous to search operators in EA. For instance, \citet{liu2025large} introduce the use of LLMs as evolutionary operators to tackle multi-objective problems. This method involves breaking down a multi-objective problem into simpler single-objective tasks, with LLMs acting as black-box search operators for each sub-problem to suggest new solutions. In a related study, \citet{liu2024largelmea} explore the integration of LLMs within EAs, not just for generating solutions but also for guiding selection, crossover, and mutation processes. Meanwhile, \citet{brahmachary2024large} propose a new population-based evolutionary framework that includes both exploration and exploitation pools, with solutions being exchanged during the optimization process and LLMs generating solutions for both pools. 

Differing from direct solution generation, \citet{lange2024large} investigate the use of LLMs in designing evolution strategies, introducing a new prompting strategy to enhance the mean statistic in their EvoLLM method, which shows superior performance over baseline algorithms in synthetic black-box optimization functions and neuroevolution tasks. They also demonstrate that fine-tuning LLMs with data from teacher algorithms can further improve the performance of EvoLLM. \citet{custode2024investigation} present a preliminary study that uses LLMs to automate hyperparameter selection by analyzing optimization logs and providing real-time recommendations. Moreover, \citet{xu2025autoep} adopt LLMs to adaptively adjust the hyperparameter for metaheuristic algorithms.

Beyond traditional optimization tasks, LLMaO has been widely adopted in prompt engineering for LLMs, a process often referred to as ``automatic prompt optimization''~\citep{zhou2022large}. These methods primarily involve iterative refinement of prompts by LLMs to improve their effectiveness for specific models (typically LLMs). Techniques include resampling-based strategies, where LLMs generate variations of original prompts while maintaining semantic similarity~\citep{wang2023promptagent}, and reflection-based strategies, where LLMs optimize by analyzing and learning from previous prompt iterations or errors~\citep{guo2023connecting}, have been explored. \citet{ma2024large} note that LLM optimizers often struggle to accurately identify the root causes of errors during the reflection process, influenced by their pre-existing knowledge rather than an objective analysis of mistakes. To address these issues, they propose a new approach termed ``automatic behavior optimization'', aimed at directly and more effectively controlling the behavior of target models. \citet{liu2025language} introduce RSBench, a benchmark set specifically for the task of evaluating LLM-based evolutionary algorithms in optimizing recommendation prompts in recommender systems.

\paragraph{Discussion:} Traditional optimizers rely on numerical or symbolic update rules. LLMaO introduces a language-conditioned optimization process, where LLMs propose and refine candidate solutions by reasoning over problem descriptions, instance context, and past trajectories rather than following fixed mathematical updates.
\begin{itemize}
    \item Advantages: 1) Leverages pre-trained domain knowledge and natural-language context. For example, in the TSP, conventional operators (e.g., 2-opt~\citep{laporte1992traveling}) improve routes based on hard-coded heuristics that do not exploit textual instance information. In contrast, LLM-based optimizers can condition on problem descriptions, constraints, previously evaluated tours and history information to produce informed improvements. \citet{liu2024largelmea} integrate LLMs as evolutionary optimizers; given task descriptions, existing solutions, and stepwise traces, the LLM performs multi-step search and generates new candidates. 2) Enables adaptive refinement without handcrafted rules. \citet{yang2024large} show that LLMs can infer optimization directions from prompt-supplied trajectories on small-scale problems, and \citet{nie2024importance} demonstrate that natural-language feedback during search further enhances LLM-driven optimization. 
    \item Limitations: 1) Limited interpretability and theoretical guarantees due to the black-box nature. Early efforts approximate LLM behavior with linear operators \citep{liu2025large} and analyze convergence under restricted settings \citep{lee2025convergence}, but general frameworks remain open. 2) Sensitivity to prompts and domain priors. Competitive results often require careful prompt design and alignment with the LLM’s training distribution \citep{yang2024large}. 3) Computational and token cost at scale. Most evaluations focus on small instances \citep{liu2024largelmea,yang2024large}; scaling to large problems is challenging due to long inputs/outputs, inference latency, and costs \citep{liu2025large}.
\end{itemize}

\subsection{LLM as Predictor (LLMaP)}

\begin{figure}[htbp]
    \centering
    \includegraphics[width=0.9\linewidth]{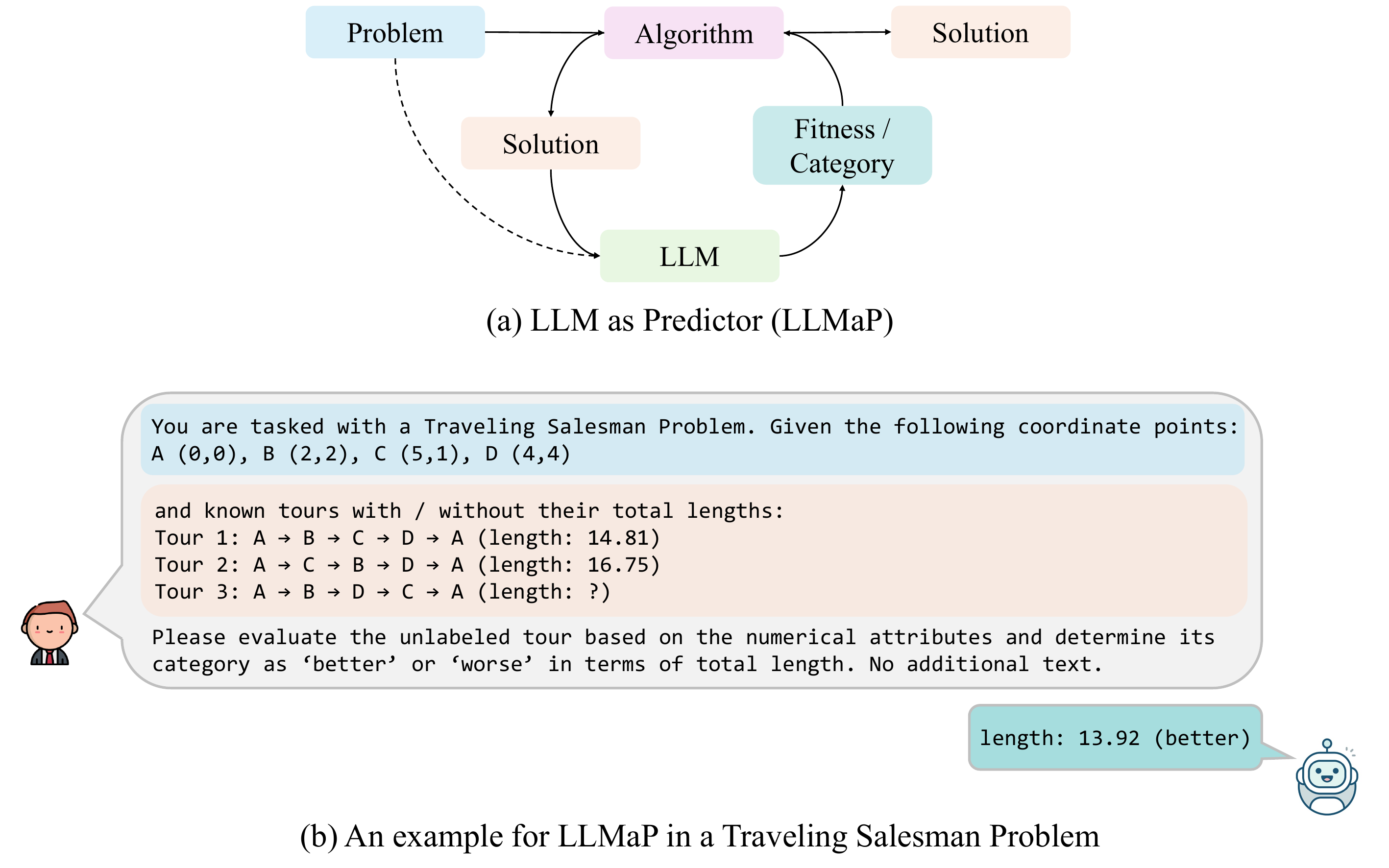}
    \Description{}
    \caption{(a) Large Language Models as Predictors (LLMaP). LLMs are Utilized Iteratively in Algorithms to Predict a Solution's Outcomes or Responses. (b) An Example for LLMaP on Traveling Salesman Problem.}
    \label{fig:llmap}
\end{figure}

LLMaP utilizes LLMs as surrogate models (Fig.~\ref{fig:llmap}) to predict the outcomes or responses of solutions, operating in either a classification or regression context~\citep{hao2024large}. Fig.~\ref{fig:llmap} (b) provides an example on a simple TSP instance, where the LLM is given the task description and a set of existing evaluated tours along with their corresponding lengths. The LLM is then instructed to evaluate a new tour, predicting its length and performance (whether it improves upon the existing tours). Note that while the length of a TSP tour is easily calculated and serves here as a straightforward illustration, LLMaP is designed for tasks where evaluations are typically expensive or difficult to obtain.


The majority of LLMaP works use LLMs as pre-trained models as a regression model to predict solution scores. For instance, LLMs have been used as performance predictors for deep neural network architectures by \citet{jawahar2024llm}. It offers a cost-effective alternative for performance estimation in neural architecture search. \citet{zhang2024can} introduce LINVIT, an algorithm that incorporates guidance from LLMs as a regularization factor in value-based RL to improve sample efficiency. Science discovery is another domain that LLMaP has commonly investigated. For example, \citet{li2023codonbert} introduce CodonBERT for sequence optimization of mRNA-based vaccines and therapeutics. CodonBERT uses codons as inputs and is trained on over 10 million mRNA sequences from various organisms. \citet{soares2024capturing} demonstrate the use of LLMs in predicting the performance of battery electrolytes. Other applications include employing LLMs to determine the fame score of celebrities to predict the box office performance of projects in the motion pictures industry~\citep{alipour2024data} and adopting LLMs to score the video question answering by using detailed video captions as content proxies~\citep{zhang2024direct}.

For classification, \citet{hao2024large} introduce LAEA, which employs LLMs as surrogate models within evolutionary algorithms for both regression and classification, eliminating the need for costly model training. In another study, \citet{chen2023label} develope a label-free node classification method that leverages LLMs to annotate nodes. These annotations are subsequently used to train graph neural networks, resulting in enhanced performance. Moving beyond binary classification, \citet{bhambri2024efficient} utilize LLMs to predict discrete actions for constructing reward shaping functions in Reinforcement Learning (RL). Their method demonstrate effectiveness within the BabyAI environment, showcasing the versatility of LLMs in various settings. \citet{wang2024resllm} explore the use of LLMs in federated search, applying them in a zero-shot setting to effectively select resources. This approach highlights the potential of LLMs in improving resource selection without prior explicit training on specific tasks.


\paragraph{Discussion:} 
Conventional surrogate models (e.g., Gaussian processes (GPs) or neural regressors~\citep{jin2018data}) rely on structured features and explicit training. In contrast, LLMaP leverages pre-trained LLMs as semantic surrogates that can interpret textual and multimodal context when predicting outcomes.
\begin{itemize}
    \item \textbf{Advantages:} 1) Effective in data-scarce or concept-driven tasks due to embedded general knowledge~\citep{egami2024using}. For example, \citet{wong2024generative} employ a multimodal LLM to score car shapes, accelerating early-stage design by filtering out poor candidates before costly simulation signals that conventional surrogates built on numeric features struggle to capture. \citet{ge2025mora} adopt LLMs to extract deeper interest preferences from the user’s behaviour and interaction history to dynamically adjust the prediction of user’s rating of items in the recommendation algorithm. 2) Requires little or no retraining, reducing computational cost. As discussed by \citet{hao2024large}, conventional surrogates often require repeated rebuilding or fine-tuning as new solutions are sampled, which is expensive for large-scale cases. LLMs can be used zero-shot or with lightweight in-context examples and can use flexible descriptors (text, code, images), thereby avoiding tedious feature engineering.
    \item  \textbf{Limitations:} 1) Quantitative precision is often inferior to specialized regression models. In the evaluation by \citet{hao2024large}, LLMs, though flexible, often achieve lower accuracy compared to GP models trained directly on the target data. \citet{xie2025large} propose hybrid schemes that combine LLMs with conventional surrogates: instead of predicting scores directly, the LLM selects which specialized surrogate to use, offering a pragmatic compromise between performance and efficiency. 2) Outputs are sensitive to prompt design and phrasing. For example, \citet{taboada2025ontology} use LLMs for ontology alignment and find that simple prompts limit performance, whereas incorporating contextual ontology information improves matching accuracy. This sensitively introduces additional costs for prompt refinement, which relies on domain knowledge.
\end{itemize}

\subsection{LLM as Extractor (LLMaE)}

\begin{figure}[htbp]
    \centering
    \includegraphics[width=0.9\linewidth]{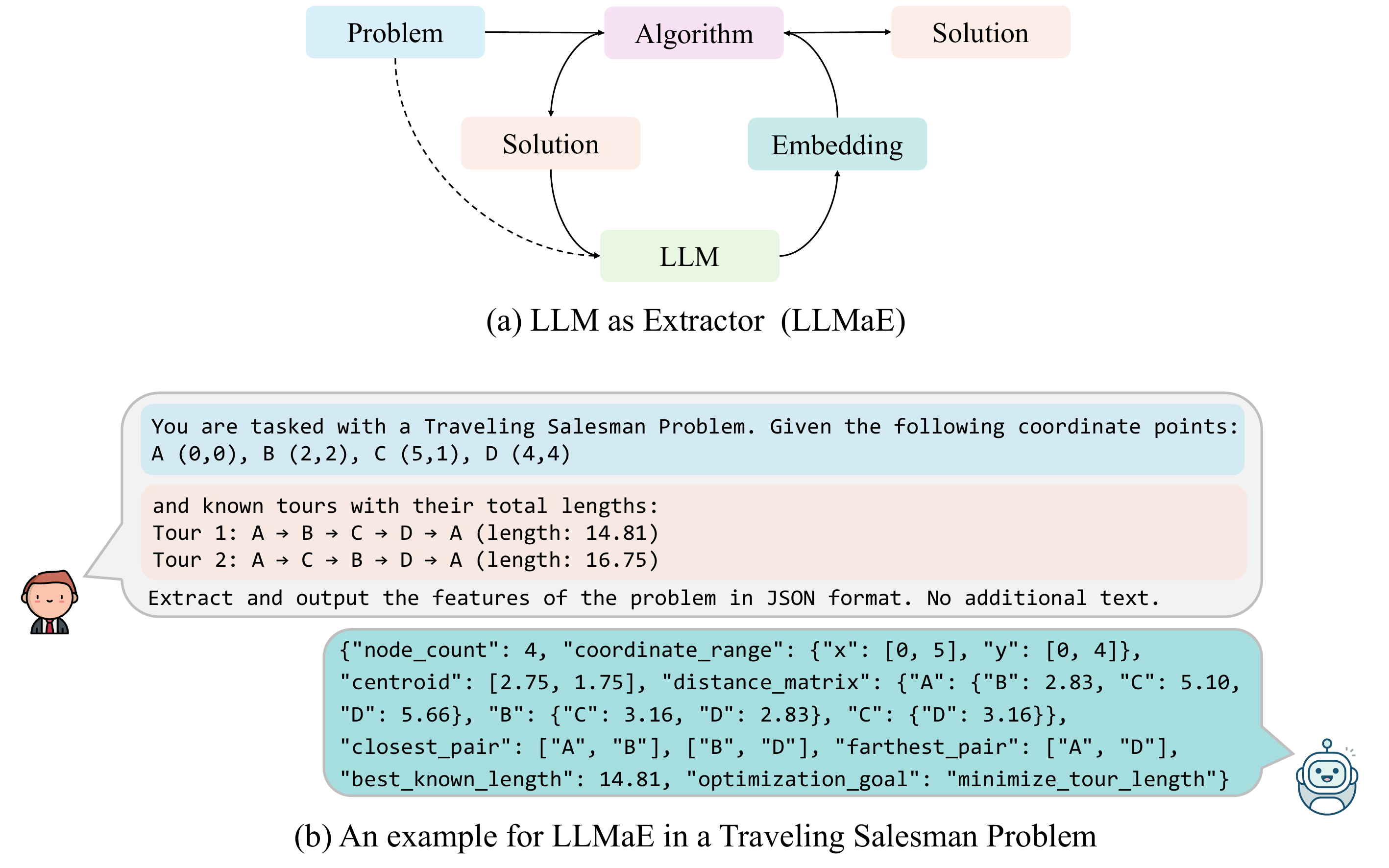}
    \Description{}
    \caption{(a) Large Language Models as Extractors (LLMaE). LLMs are Employed to Extract Features or Specific Knowledge from Target Problem and/or Algorithms. (b) An Example for LLMaE on Traveling Salesman Problem.}
    \label{fig:llmae}
\end{figure}

LLMaE leverages LLMs to mine and extract embedded features or specific knowledge from target problems and/or algorithms, which are then used to enhance algorithm-based problem solving (Fig.~\ref{fig:llmae}). As shown in Fig.~\ref{fig:llmae} (b), when provided with the task description and tours for a TSP instance, LLMs can be instructed to extract key features, such as node count, which can then be utilized in algorithm design. These features are typically defined by experts or learned implicitly using conventional machine learning models.


Beyond embedding-based feature extraction, LLMs excel in text comprehension and knowledge extraction, allowing them to discern subtle patterns and relationships within the data that might not be evident through conventional feature extraction methods. For example, \citet{wu2024large} utilize LLMs to extract high-dimensional algorithm representations by comprehending code text. These representations are combined with problem representations to determine the most suitable algorithm for a specific problem. \citet{du2024mixture} propose a mixture-of-experts framework augmented with LLMs to optimize various wireless user tasks. The LLM is used to analyze user objectives and constraints, thus selecting specialized experts, and weighing decisions from the experts, reducing the need for training new models for each unique optimization problem. Additionally, \citet{memduhouglu2024enriching} use LLM to enhance the classification of urban building functions by interpreting OpenStreetMap tags and integrating them with physical and spatial metrics. Traditional techniques, which have previously struggled with semantic ambiguities, are outperformed by LLMs due to their superior ability to capture broader language contexts. Beyond extracting problem features, LLMs are employed to mine relevant knowledge to inform and enhance algorithm design. In HiFo-Prompt \citep{chen2025hifo}, for example, knowledge is distilled into reusable design principles that guide the design process. 


Typical feature extraction relies on statistical learning or dimensionality reduction from structured data. LLMaE instead is able to perform semantic extraction, deriving contextual features and domain knowledge from unstructured sources such as text, code, or documentation.
\begin{itemize}
    \item \textbf{Advantages:} 1) It generates concept-aware embeddings that integrate both linguistic and symbolic meaning. For instance, \citet{xu2025evospeak} use LLMs to analyze existing heuristic structures and extract underlying design principles and domain-relevant insights, thereby providing a warm start to enhance the quality of designed heuristics. 2) It combines textual, spatial, and numeric features into a unified representation. For example, \citep{wu2024large} use LLMs to extract algorithm features for algorithm selection. Unlike traditional approaches, such as machine learning prediction models, both text descriptions and code implementations can be leveraged.
    \item \textbf{Limitations:} 1) There is a risk of hallucinated or misinterpreted features in specialized domains. For instance, \citep{hu2025discovering} report information loss and misinterpretation when using multi-modal LLMs for designing algorithms for agents. 2) The validation and interpretability of embeddings remain challenging. \citet{jiang2024bridging} use LLM embeddings to solve the vehicle routing problem. While the extracted information proves beneficial in ablation studies, providing a clear interpretation and analysis of the embedding remains difficult.
\end{itemize}


\subsection{LLM as Designer (LLMaD)}

\begin{figure}[htbp]
    \centering
    \includegraphics[width=0.9\linewidth]{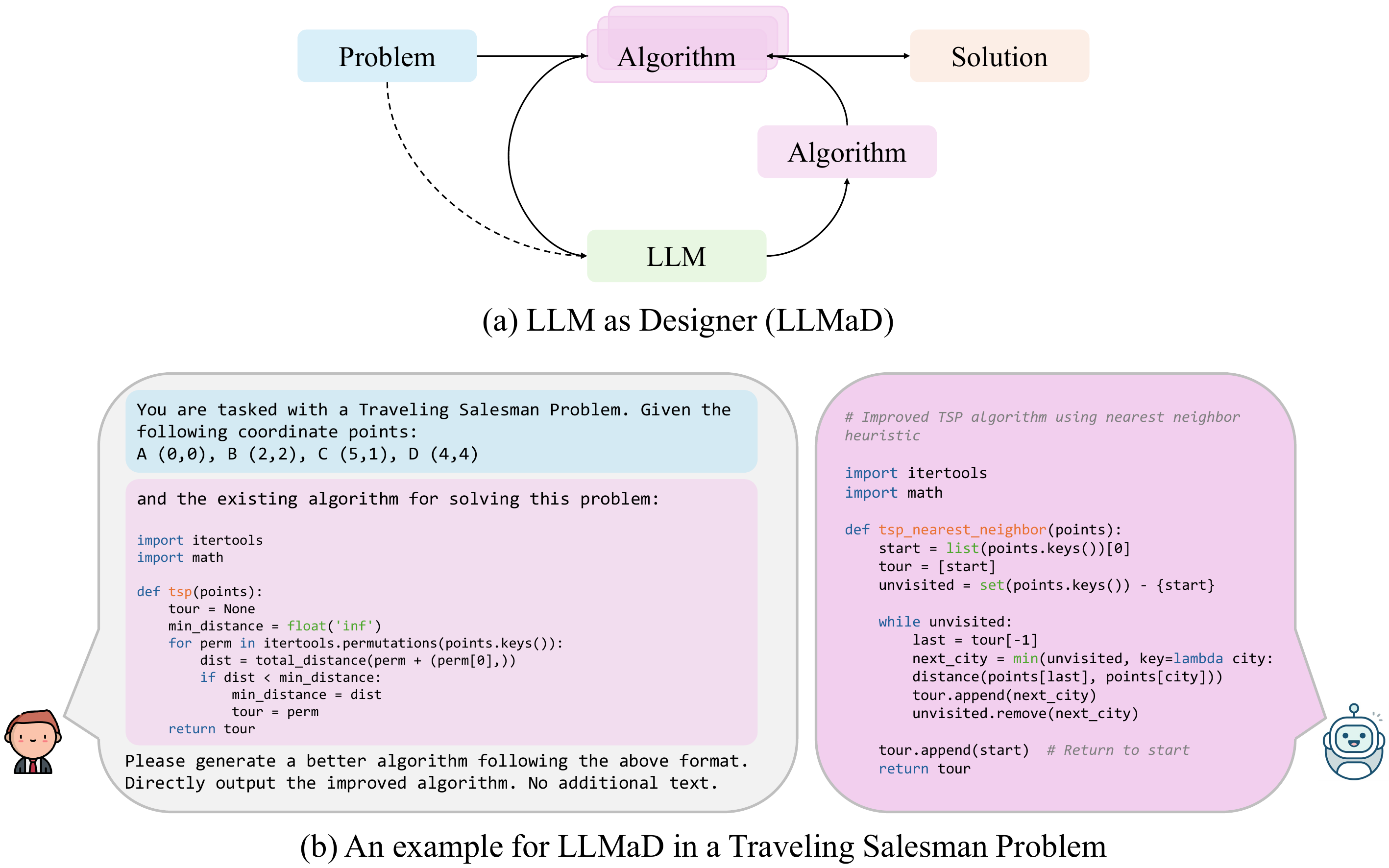}
    \Description{}
    \caption{(a) Large Language Models as Designers (LLMaD). LLMs are Used to Directly Create the Entire Algorithms or Specific Components. (b) An Example for LLMaD on Traveling Salesman Problem.}
    \label{fig:llmad}
\end{figure}

LLMaD directly creates algorithms or specific components (Fig.~\ref{fig:llmad}). LLMs can generate heuristics~\citep{liu2024evolution}, write code snippets~\citep{hemberg2024evolving}, and formulate functions~\citep{romera2024mathematical} that integrate seamlessly into algorithmic systems, or even design the entire algorithms~\citep{van2024llamea,novikov2025alphaevolve}. Fig.~\ref{fig:llmad} demonstrates this process using a TSP instance. Unlike other paradigms, LLMaD directly produces new algorithms or algorithm components based on a task description and an existing algorithm implementation.



Function design is among the early applications of LLMaD. Eureka~\citep{ma2024eureka} leverages the capabilities of LLMs in code-writing, and in-context learning to evolve and optimize reward functions for RL. It can generate reward functions without specific prompts or predefined templates, achieving better performance than rewards designed by human experts. Similarly, Auto MC-Reward~\citep{li2024auto} utilizes LLMs to automatically design dense reward functions for RL agents in environments with sparse rewards. The three key components of Auto MC-Reward work together to iteratively refine the reward function based on feedback from the agent's interactions with the environment. Through this iterative process, the agent is able to learn complex tasks more efficiently, as demonstrated in experiments in Minecraft. Moreover, FunSearch~\citep{romera2024mathematical} adopts LLMs for function generation in an evolutionary framework with a multi-island population management. It demonstrates promising results on both mathematical problems and combinatorial optimization problems.


EoH~\citep{liu2024evolution,liu2023algorithm} presents an early attempt to adopt the LLM as a designer for Automated Heuristic Design (AHD). It uses both heuristic ideas and code implementations to represent heuristics and adopts LLM in an evolutionary framework to create, combine, and revise the heuristics. While it is original proposed for heuristic design, it has been applied on different algorithm design tasks including combinatorial optimization problems~\citep{liu2024evolution,yao2024multi,liu2025eoh}, Bayesian optimization~\citep{yao2024evolve}, image adversary attack~\citep{guo2024autoda}, and edge server task scheduling~\citep{wang2024ts}, among others. Moreover,  LLaMEA~\citep{van2024llamea,van2024loop} develops an iterative framework to generate, mutate, and select algorithms based on performance metrics and runtime evaluations. The automatically designed algorithms outperform state-of-the-art optimization algorithms on some benchmark instances. ReEvo~\citep{ye2024reevo} introduces an evolutionary framework with both short and long-term reflections, which provides a search direction to explore the heuristic space. HSEvo \citep{dat2025hsevo} and PartEvo \citep{hu2025partition} integrate different diversity control strategies in evolutionary search framework to enhance the search. In addition to evolutionary search framework, recent attempts also adopt other frameworks such as large neighborhood search~\citep{xie2025llm} and Monte Carlo Tree Search (MCTS)~\citep{zhengmonte} for effective explore the algorithm space. Unlike previous studies that focus on optimizing a single performance criterion, MEoH~\citep{yao2024multi} considers multiple performance metrics, including optimality and efficiency, and seeks a set of trade-off algorithms in a single run in a multi-objective evolutionary framework. A dominance-dissimilarity score is designed for effectively searching the complex algorithm space.

LLM-based agent design has also gained much attention. For example, ADAS~\citep{hu2024automated} proposes an automated design of agentic systems, which aims to automatically generate powerful agentic system designs by using meta agents that program new agents. They present a novel algorithm, meta agent search, which iteratively creates new agents from an archive of previous designs, demonstrating through experiments that these agents can outperform state-of-the-art hand-designed agents across various domains. Further studies on LLM-based agent systems are discussed in \citep{sun2024llm} and \citep{liu2024agentlite}.


\paragraph{Discussion:} Different from existing approaches, such as human-crafted heuristics, AutoML, and genetic programming, search within explicit procedural representations. LLMaD advances this by using LLMs for language-based algorithm synthesis, generating algorithms or components via natural-language reasoning and code generation.
\begin{itemize}
    \item \textbf{Advantages:} 1) Accelerates algorithm design by reducing human intervention and implementation effort. LLMaD often operates zero-shot or with lightweight iterative prompting; the outputs are executable text/code that can be inspected, profiled, and reused, while achieving competitive efficiency~\citep{liu2024evolution}. 2) Supports iterative self-improvement when embedded in feedback or evolutionary loops, expanding the creativity and diversity of algorithmic solutions~\citep{zhang2025darwin,van2025code}. For instance, \citet{zhang2025darwin} propose a tree-based search framework that uses LLMs to design diverse, high-quality agents in an open-ended manner.
    \item \textbf{Limitations:} 1) LLMs struggle with synthesizing complete, state-of-the-art algorithms for complex tasks; most studies target specific components (e.g., heuristics, reward functions, code snippets) rather than the complete solvers~\citep{novikov2025alphaevolve}. Challenges around correctness, robustness, and runtime guarantees persist. 2) Algorithm design is domain-specific, and standalone LLMs typically lack sufficient task-specific knowledge, leading to subpar performance~\citep{zhang2024understanding}. Effective LLMaD systems therefore depend on search frameworks that iteratively test, execute, and refine designs while interacting with the environment and tools~\citep{zhang2024understanding,van2025blade}.
\end{itemize}

\begin{figure}
    \centering
    \includegraphics[width=0.95\linewidth]{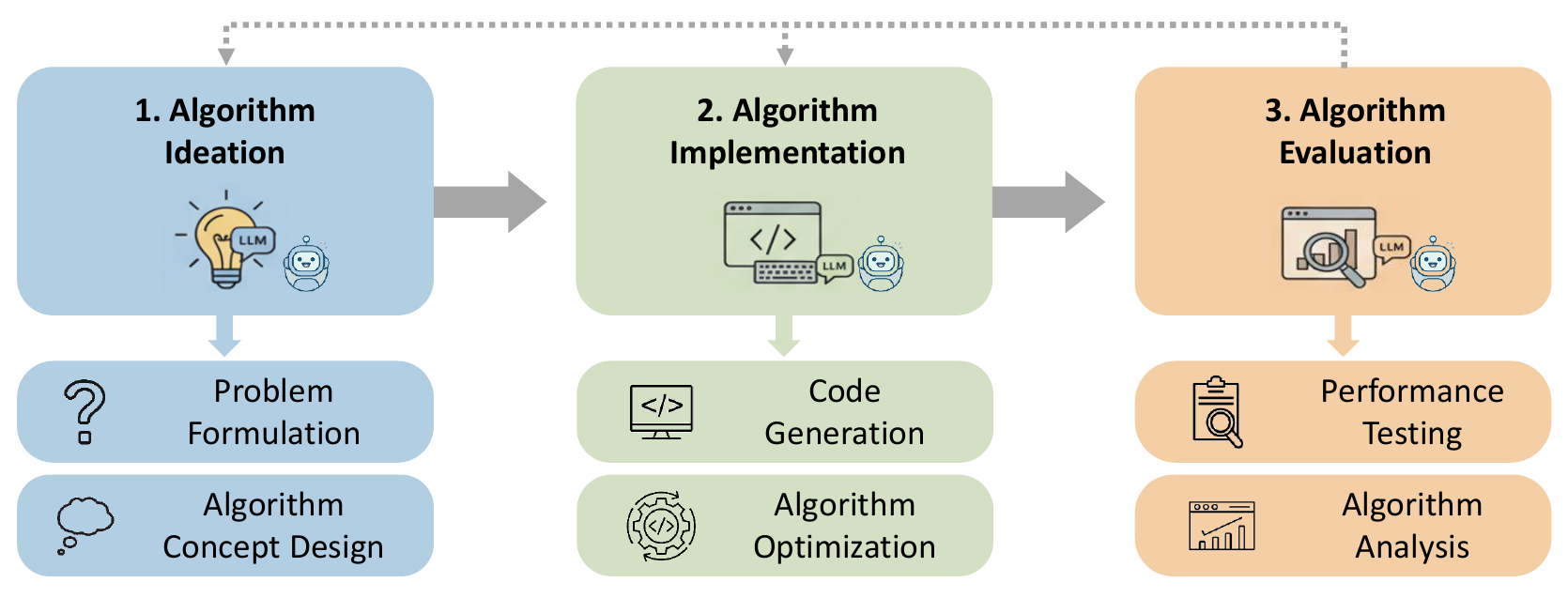}
    \caption{The using of LLMs in Different Stages of Algorithm Design: 1) Algorithm Ideation, 2) Algorithm Implementation, and 3) Algorithm Evaluation.}
    \label{fig:placeholder}
\end{figure}

\section{Stages of LLM‑Assisted Algorithm Design}~\label{sec4:stage}

Algorithm design is a multi-stage process where LLMs can provide targeted assistance at distinct phases. We identify three key stages in this process: 1) algorithm ideation, 2) algorithm implementation, and 3) algorithm evaluation. This section organizes the existing literature along these three stages, summarizing the typical artifacts, recent progresses and challenges.

\subsection{Stage I: Algorithm Ideation}
The ideation stage encompasses both problem formulation and the generation of novel algorithmic concepts.

\paragraph{Problem Formulation.}
While not a direct step in creating an algorithm, modeling the target problem is a crucial precursor to its design, and LLMs are increasingly used to automate this process. Early systems, such as OptiMUS, guide users from natural language descriptions to formal linear programs, producing executable solver \citep{ahmaditeshnizi2024optimus}. To mitigate the privacy and controllability concerns associated with proprietary APIs, ORLM combines an instruction-following recipe (OR-Instruct) with open-source backbone models, reporting strong results on benchmarks like NL4Opt and the new IndustryOR benchmark \citep{tang2024orlm,ramamonjison2023nl4opt}. Complementary fine-tuning strategies, such as those in LM4OPT, demonstrate that even modest-sized models can be specialized to reliably map domain-specific text to optimization modeling \citep{ahmed2024lm4opt}. In the context of interactive decision support, DOCP integrates user feedback to progressively assemble problem-specific models before invoking solvers \citep{wasserkrug2024large}. For a comprehensive overview, \citet{xiao2025survey} provide an in-depth survey of LLMs for optimization modeling.

\paragraph{Algorithm Concept Design.}
LLMs have shown promise in generating new algorithmic ideas across various domains \citep{si2024can}. For instance, \citet{pluhacek2023leveraging} use LLMs to analyze and decompose well-performing algorithms to propose hybrid algorithms that combine their complementary strengths. While this work demonstrated that novel algorithmic ideas could be designed, it did not implement them for specific algorithm design tasks. EoH presents an early attempt to co-design algorithm concepts in natural language alongside their executable code implementations \citep{liu2024evolution}. Different strategies have been explored to enhance this co-design search process, including reasoning over algorithmic ideas during design \citep{bomer2025leveraging,ling2025complex} and controlling the diversity to foster innovation \citep{yao2025evolution}. 

However, how to effectively measure the quality and novelty of an algorithmic idea remains an open question~\citep{liu2025fitness}. Moreover, the concept descriptions are often highly abstract, which lacks a strict mapping to a detailed implementation, therefore making it hard to effectively guide algorithm implementation \citep{gurkan2025lear}.

\subsection{Stage II: Algorithm Implementation}
The implementation stage spans from initial code synthesis to subsequent performance optimization.

\paragraph{Code Generation.}
Given an algorithmic idea or a natural language description, LLMs can be instructed to generate the corresponding code implementation~\citep{liu2024evolution,yao2024evolve,zhengmonte}. Some works bypass the explicit ideation stage and directly use code as the primary representation of an algorithm~\citep{romera2024mathematical,ye2024reevo}. Instead of generating code from structure, ADAS employs meta-agents to program new agents from an archive of prior designs, thereby automating the construction of multi-component pipelines \citep{hu2024automated}. Initially, code generation focused on individual algorithmic components, reflecting the limited capabilities of early LLMs and simpler design pipelines~\citep{romera2024mathematical,liu2024evolution}. More recently, this has been extended to generating entire algorithms~\citep{van2024llamea}. To handle large-scale code, \citet{novikov2025alphaevolve} use ``diff blocks'' to highlight specific sections, enabling the model to generate or revise partial code instead of regenerating the entire program. To improve reliability at scale, LLaMoCo introduces instruction tuning tailored to code generation for optimization tasks, reducing the reliance on expert-crafted prompts \citep{ma2024llamoco}. 

Despite these advances, the generation of entire, complex algorithms remains a significant challenge. First, the reliable generation of long programs is a difficult problem in itself~\citep{jiang2024survey}. Second, it is challenging to ensure that a complicated implementation correctly instantiates the intended algorithmic idea and is effective for the target task. Addressing this requires not only powerful LLMs but also sophisticated frameworks, such as multi-agent systems~\citep{Ishibashi2024selforganized}, to manage the complexity.
\\
\paragraph{Algorithm Optimization.} 
A common pattern for algorithmic code optimization is closed-loop refinement, where execution feedback informs subsequent code edits and integration choices~\citep{zhang2024understanding}. Evolutionary search is the most prevalent framework for this purpose, maintaining a population of algorithms that are progressively refined by an LLM~\citep{zhang2024understanding}. Variants of this simple evolutionary approach have emerged. For instance, ReEvo \citep{ye2024reevo} combines short- and long-term reflection with information on history evaluated results to guide algorithm optimization, while HSEvo \citep{dat2025hsevo} adopts diversity control strategies with two population diversity measurements to enhance the optimization process. Recently, non-evolutionary search frameworks, such as large neighborhood search and Monte Carlo tree search~\citep{zhengmonte}, have also been demonstrated to be effective. Moreover, instead of prompting LLMs to optimize algorithm implementations, recent attempts have explored fine-tuning LLMs, in both offline~\citep{liu2025fine} and online manners~\citep{huang2025calm}, to learn a preference for generating better-performing algorithms during algorithm optimization for specific design tasks.

A significant open challenge is ensuring that the optimized algorithms generalize robustly. An algorithm may be overfitted to the specific benchmarks or problem instances used during the optimization loop, failing to perform well on unseen instances \citep{sim2025beyond} or under different conditions \citep{shi2025generalizable}.

\subsection{Stage III: Algorithm Evaluation}
The final stage, algorithm evaluation, involves assessing the performance and analyzing the behavior of the designed algorithms.
\paragraph{Performance Testing.}
To ensure comparability and robustness in LLM-driven algorithm design, recent efforts have focused on establishing standardized benchmarks. Notable examples include CO-Bench and HeuriGym, which are tailored for structured combinatorial optimization tasks \citep{sun2025co,chen2025heurigym}. In addition to combinatorial optimization, LLM4AD \citep{liu2024llm4ad} provides a unified evaluation platform that spans a broader range of domains, including optimization, machine learning, and scientific discovery \citep{liu2024llm4ad}. Beyond static benchmarks for testing, some approaches leverage LLMs to dynamically refine the evaluation process itself. For instance, \citet{li2025cocoevo} propose a method to co-evolve algorithms and their test instances, aiming to generate more effective test cases that better guide the algorithm design process. Similarly, \citet{duan2025ealg} introduce a mutation-based adversarial approach that dynamically evolves instance generation procedures to create increasingly difficult problems, thereby enhancing the generalization performance of the designed algorithms.

Moving beyond a single performance score, a deeper evaluation should analyze the algorithm's behavior, including its search trajectory, robustness across different problem distributions, convergence properties, and computational efficiency~\citep{yao2024multi}. LLMs are uniquely positioned to automate this in-depth analysis. Just as a human expert would, they can interpret search patterns, identify failure modes, and diagnose the root causes of poor performance to provide actionable, human-like insights to directly enhance the design process~\citep{yu2024deep}.

\paragraph{Algorithm Analysis.}
In addition to generation, LLMs are also being employed to analyze algorithmic outcomes and explain their behavior. \citet{d2024exploring} empirically analyze and compare the quality of explanations provided by three different LLMs for seven state-of-the-art quantum algorithms. Their findings indicate that while the explanations were consistent across multiple iterations, there remains a significant gap for improvement in quality, and the results were highly sensitive to the specific prompts used. In a different application, \citet{chacon2024large} integrate LLMs into STNWeb, an algorithm analysis tool. The LLM generates extensive written reports, complemented by automatically generated plots, which enhances the user experience and lowers the barrier to adoption for the broader research community.

The central challenge for LLM-based algorithm analysis is the gap between descriptive summarization and causal, principled understanding. Current methods can generate consistent and well-written reports, but these often remain surface-level descriptions of behavior rather than deep explanations rooted in algorithmic theory~\citep{chacon2024large}. The analysis quality is also highly sensitive to prompt engineering, making it unreliable and not yet a trustworthy source of foundational insight. For the field to advance, LLMs must move beyond post-hoc commentary to providing causal reasoning about why an algorithm performs as it does. For instance, by explain its convergence properties, time complexity, or failure modes based on its structural components~\citep{kleinberg2006algorithm,van2025code,liu2025fitness}.

\section{Applications}\label{sec5:appli}

\subsection{Optimization Algorithms}\label{subsec:dis_opt}
In this subsection, we delve into the applications of LLMs in designing optimization algorithms. We categorize the existing literature into combinatorial optimization and continuous optimization. Then we proceed to compare the various roles played by LLMs, and the specific problems or tasks to which they are applied. The comparative analysis is summarized in Table \ref{tab:optimization application}, where we list the names of the frameworks or methods proposed by the authors. For studies that do not explicitly name their methods, we assign appropriate designations in our article and denote them with asterisks for easy reference (e.g., MH-LLM* for \citep{sartori2024metaheuristics}).

\subsubsection{Combinatorial Optimization}
In the domain of Combinatorial Optimization (CO), automated algorithm heuristics design has been a significant area of interest for a long time. The Traveling Salesman Problem (TSP) stands out as one of the most renowned CO problems, involving the quest for the shortest route to visit all specified locations exactly once and return to the starting point. Some recent work leverages LLMs to evolve algorithms within evolutionary computation framework, such as EoH \citep{liu2024evolution} and ReEvo \citep{ye2024reevo}. Differently, OPRO \citep{yang2024large} employs LLMs as optimizers with a proposed meta-prompt, in which the solution-score pairs with task descriptions are added in each optimization step. Additionally, LMEA \citep{liu2024largelmea} investigates the utilization of LLMs as evolutionary combinatorial optimizers for generating offspring solutions, wherein a self-adaptation mechanism is introduced to balance exploration and exploitation. The Capacitated Vehicle Routing Problem (CVRP) extends the TSP by introducing constraints related to vehicle capacity. To address this challenge, MLLM \citep{huang2024multimodal} devises a multi-modal LLM-based framework with textual and visual inputs to enhance optimization performance. In addition to routing problems, other combinatorial optimization problems that have also been investigated include cap set~\citep{romera2024mathematical}, bin packing~\citep{liu2024evolution}, flow shop scheduling~\citep{liu2024evolution}, hybrid job shop scheduling~\citep{li2025llm} and social networks problems~\citep{sartori2024metaheuristics}.

\subsubsection{Continuous Optimization}
For single-objective optimization problems with continuous variables, LLaMEA \citep{van2024llamea} utilizes LLMs to automate the evolution of algorithm design. It demonstrates the effectiveness in generating new metaphor-based optimization algorithms on BBOB benchmark \citep{hansen2010black} within IOHexperimenter benchmarking tool \citep{de2024iohexperimenter}, which supports evaluating the quality of the generated algorithms and also provides feedback to the LLM during evolution. Instead of creating new algorithms, EvolLLM \citep{lange2024large} introduces a prompt strategy that enables LLM-based optimization to act as an Evolution Strategy (ES) and showcases robust performance on synthetic BBOB functions and neuroevolution tasks. OPRO \citep{yang2024large} illustrates that LLMs can effectively capture optimization directions for linear regression problems by leveraging the past optimization trajectory from the meta-prompt. Additionally, LEO \citep{brahmachary2024large} devises an explore-exploit policy using LLMs for solution generation, the method has been tested in both benchmark functions as well as industrial engineering problems. Different with directly employing LLMs for generating solutions, LAEA \citep{hao2024large} introduces LLM-based surrogate models for both regression and classification tasks and has been validated on 2D test functions using nine mainstream LLMs.

Beyond a single objective, there are multiple competing objectives that need to be optimized simultaneously in many scenarios, forming the multi-objective optimization problems (MOPs). The goal is to identify a set of optimal solutions, referred to as Pareto optimal solution set. An initial exploration of utilizing LLMs to tackle MOPs is introduced in MOEA/D-LMO \citep{liu2025large}. Benefiting from the decomposition-based framework, the in-context learning process of LLMs is easily incorporated to generate candidate solutions for each subproblem derived from the original MOP. In the realm of large-scale MOPs, LLM-MOEA* \citep{singh2024enhancing} showcases the inferential capabilities of LLMs in multi-objective sustainable infrastructure planning problem. The study highlights the LLM's proficiency in filtering crucial decision variables, automatically analyzing the Pareto front, and providing customized inferences based on varying levels of expertise. Additionally, CMOEA-LLM \citep{wang2024large} leverages LLMs with evolutionary search operators to address the challenges of constrained MOPs and exhibits robust competitiveness in DAS-CMOP test suite \citep{fan2020difficulty}.

In various real-world applications, the cost of evaluating the objective functions can be very expensive, which greatly limits the evaluation budget in the optimization process \citep{frazier2016bayesian}. Bayesian optimization (BO) stands out as a sample-efficient method, it typically employs a surrogate model to approximate the expensive function and well-designed Acquisition Functions (AFs) to carefully select potential solutions. To facilitate the direct generation of solutions using LLMs, HPO-LLM* \citep{zhang2023using} provides LLMs with an initial set of instructions that outlines the specific dataset, model, and hyperparameters to propose recommended hyperparameters for evaluation in Hyperparameter Optimization (HPO) tasks. Furthermore, LLAMBO \citep{liu2024largellambo} incorporates LLM capabilities to enhance BO efficiency, in which three specific enhancements throughout the BO pipeline have been systematically investigated on tasks selected from HPOBench \citep{eggensperger2021hpobench}. Instead of utilizing LLMs for direct solution generation, BO-LIFT* \citep{ramos2023bayesian} utilizes predictions with uncertainties provided by a Language-Interfaced Fine-Tuning (LIFT) framework \citep{dinh2022lift} with LLMs to perform BO for catalyst optimization using natural language. EvolCAF \citep{yao2024evolve} introduces a novel paradigm to design AFs automatically for cost-aware BO. The approach showcases remarkable efficiency and discovers novel ideas not previously explored in existing literature on AF design. Similarly, FunBO \citep{aglietti2025funbo} found novel and well-performing AFs for BO by extending FunSearch \citep{romera2024mathematical}. The discovered AFs are evaluated on various synthetic and HPO benchmarks in and out of the training distribution.

\begin{table*}[htbp]
\centering
\caption{An Overview of Optimization Applications Utilizing Language Models Across Various Domains and Task. \label{tab:optimization application}}
\resizebox{0.9\textwidth}{!}{
\begin{tabular}{c|c|c|c}
\toprule
Application & Method& Role of LLM  & Specific Problems or Tasks\\
\midrule
\multirow{12}{*}{Combinatorial Optimization}
  & EoH \citep{liu2024evolution,liu2023algorithm}  & LLMaD    & TSP, Online BPP, FSSP  \\
\cmidrule{2-4}
  & ReEvo \citep{ye2024reevo} & LLMaD   & TSP, CVRP, OP, MKP, BPP, DPP   \\ 
\cmidrule{2-4}
  & OPRO \citep{yang2024large}& LLMaO    & TSP  \\
\cmidrule{2-4}
  & LMEA \citep{liu2024largelmea} & Mixed   & TSP   \\
\cmidrule{2-4}
  & MLLM \citep{huang2024multimodal}  & Mixed    & CVRP  \\
\cmidrule{2-4}
  & FunSearch \citep{romera2024mathematical}  & LLMaD    & Cap Set Problem, Online BPP\\
\cmidrule{2-4}
  & MH-LLM* \citep{sartori2024metaheuristics}  & LLMaD     & Social Networks Problem   \\
\cmidrule{2-4}
  & SolSearch \citep{sheng2025solsearch} & LLMaD    &  Satisfiability Problem \\
\cmidrule{2-4}
    & LMPSO \citep{shinohara2025large} & LLMaO  & 
  TSP \\
\cmidrule{2-4}
  & LLM-NSGA \citep{wan2025surgery}  & LLMaD & Surgery Scheduling Problem \\
\cmidrule{2-4}
  & STRCMP \citep{li2025strcmp} & LLMaD  & MILP, SAT \\
\cmidrule{2-4}
  & EvoCut \citep{yazdani2025evocut} & LLMaD & MILP
\\

\midrule
\multirow{17}{*}{Continuous Optimization}
  & LLaMEA \citep{van2024llamea}  & LLMaD   & BBOB  \\
\cmidrule{2-4}
  & EvoLLM \citep{lange2024large} & LLMaO    & BBOB, Neuroevolution  \\
\cmidrule{2-4}
  & OPRO \citep{yang2024large} & LLMaO  & Linear Regression  \\
\cmidrule{2-4}
  & LEO \citep{brahmachary2024large}  & LLMaO   & Numerical Benchmarks, Industrial Engineering Problems  \\
\cmidrule{2-4}
  & LAEA \citep{hao2024large} & LLMaP    & Ellipsoid, Rosenbrock,
Ackley, Griewank \\
\cmidrule{2-4}
  & MOEA/D-LMO \citep{liu2025large}  & LLMaO    & Mulit-objective Synthetic Functions\\
\cmidrule{2-4}
  & LLM-MOEA* \citep{singh2024enhancing} & LLMaE  & Multi-objective Sustainable Infrastructure Planning Problem   \\
\cmidrule{2-4}
  & CMOEA-LLM \citep{wang2024large}   & LLMaO    & DAS-CMOP   \\
  
\cmidrule{2-4}
  & HPO-LLM* \citep{zhang2023using} & LLMaO   & HPOBench \\
\cmidrule{2-4}

  & LLAMBO \citep{liu2024largellambo}   & Mixed  & Bayesmark, HPOBench \\
\cmidrule{2-4}

  & BO-LIFT* \citep{ramos2023bayesian} & LLMaD   & Catalyst Optimization \\
\cmidrule{2-4}

  & EvolCAF \citep{yao2024evolve} & LLMaD   & Synthetic Functions, HPO \\
\cmidrule{2-4}

  & FunBO \citep{aglietti2025funbo}  & LLMaD  & Synthetic Functions, HPO \\
\cmidrule{2-4}

  & LLM-SAEA \citep{xie2025large} & LLMaP  & Synthetic Functions \\
  
\cmidrule{2-4}
  & AwesomeDE \citep{yang2025large} & LLMaO  & Synthetic Functions \\
\cmidrule{2-4}
  & BBNM \citep{lee2025convergence}  & LLMaO  & Wireless Networks \\
\cmidrule{2-4}
  & LLM4CMO \citep{chen2025llm4cmo} & LLMaO  & Constrained Multiobjective Optimization
\\

\bottomrule
\end{tabular}
}
\vspace{-1em}
\end{table*}

\subsection{Machine Learning Systems and Data Science}




In this subsection, we investigate the applications of LLMs in the machine learning domain, focusing on their contribution to algorithmic design. These applications are summarized in Table~\ref{tab:ml_application}. 


\subsubsection{Reinforcement Learning}
Reinforcement Learning (RL) has been the de facto standard for sequential decision-making tasks, and recently, the synergy between RL and LLMs has emerged as a novel trend in the domain.
This convergence mirrors the dynamics of task planning, yet places RL at the core of its methodology. Many of the LLM4AD papers on RL is for automatically designing the reward functions~\citep{bhambri2024efficient,ma2024eureka,narin2024evolutionary}. In addition, \citet{shah2023navigation} investigate the employment of LLMs for heuristic planning to steer the search process within RL frameworks. 
\citet{zhang2024can} integrate LLMs into RL by introducing a Kullback-Leibler divergence regularization term that aligns LLM-driven policies with RL-derived policies. LLMs have also extended their reach to multi-agent RL scenarios, as shown by \citet{du2024mixture}, who illustrates their application within a Mixture-of-Experts system to direct RL models in the realm of intelligent network solutions.

\subsubsection{Neural Architecture Search}
Neural Architecture Search (NAS), which is a significant focus within the AutoML community, has been investigated in many LLM4AD papers. For example, \citet{chen2024evoprompting} have integrated LLMs with evolutionary search to successfully generate NAS code for diverse tasks. \citet{nasir2023llmatic} introduce a quality-diversity algorithm tailored to NAS, producing architectures for CIFAR-10 and NAS-bench-201 benchmarks. Moreover, \citet{morris2024llm} introduce guided evolution for the development of neural architectures and suggest the concept of the evolution of thought in algorithm design. Except for using LLM for design, \citet{jawahar2024llm} employ LLMs in predicting NAS performance, combining this approach with evolutionary search to effectively create novel network architectures. In contrast to the LLM-based architecture design and performance prediction, \citet{zhou2024design} explore the adoption of LLMs for transferring design principles to narrow and guide the search space.

\subsubsection{Prompt Tuning}
Prompt tuning aims to identify the most effective task prompt to enhance the performance of the LLM on a specific task dataset. Despite of requiring specialized training for each specific task, traditional discrete or continuous approaches \citep{li2021prefix} typically necessitate access to the logits or internal states of LLMs, which may not be applicable when the LLM can only be accessed through an API. To address these issues, recent works propose to model the optimization problem in natural language with LLMs as prompts. APE \citep{zhou2022large} utilizes the LLM as an inference model to generate instruction candidates directly based on a small set of demonstrations in the form of input-output pairs. This approach has demonstrated human-level performance on various tasks, including Instruction Induction \citep{honovich2022instruction} and Big-Bench Hard (BBH) \citep{suzgun2022challenging}. OPRO \citep{yang2024large} enables the LLM as an optimizer to gradually generate new prompts based on the full optimization trajectory, the optimizer prompt showcases significant improvement compared with human-designed prompts on BBH and GSM8K \citep{cobbe2021training}. Inspired by the numerical gradient descent method, APO \citep{pryzant2023automatic} conducts textual “gradient descent” by identifying the current prompts’ flaws and adjusting the prompt in the opposite semantic direction of the gradient. Similar practices are also found in the gradient-inspired LLM-based prompt optimizer named GPO \citep{tang2024unleashing}, as well as the collaborative optimization approach \citep{guo2024two} integrating a gradient-based optimizer and an LLM-based optimizer. Differently, \citet{guo2023connecting} introduce a discrete prompt tuning framework named EvoPrompt that prompts LLM to act like evolutionary operators in generating new candidate prompts, harnessing the benefits of evolutionary algorithms that strike a good balance between exploration and exploitation. StrategyLLM \citep{gao2023strategyllm} integrates four LLM-based agents—strategy generator, executor, optimizer, and evaluator—to collaboratively induce and deduce reasoning. This method generates more generalizable and consistent few-shot prompts than CoT prompting techniques.

\subsubsection{Graph Learning}
Graph learning is another application with the advancing capabilities of LLMs in symbolic reasoning and graph processing. For example, \citet{chen2023label} apply LLMs to the task of labeling in Text-Attributed Graphs (TAGs), capitalizing on the language task proficiency of LLMs. Both \citet{mao2024identify} and \citet{chen2024hin} adopt LLMs in an evolutionary framework for designing functions. The former evolves heuristic code functions to identify critical nodes in a graph while the latter identifies meta-structures within heterogeneous information networks to enhance their interpretability. Moreover, knowledge graphs have also seen substantial benefits from the application of LLMs. \citet{zhou2024autoalign} introduce AutoAlign, a method that employs LLMs to semantically align entities across different knowledge graphs, and \citet{feng2023knowledge} develop the knowledge search language to effectively conduct searches within knowledge graphs.

\subsubsection{Dataset Labeling}
LLMs have been used for mining semantic and multi-modal information from datasets. LLMs are employed to train interpretable classifiers to extract attributes from images~\citep{chiquier2024evolving} and to generate label functions for weakly supervised learning~\citep{guan2023can}.

\begin{table*}[htbp]
\centering
\caption{An Overview of Machine Learning Applications Utilizing Language Models Across Various Domains and Tasks.\label{tab:ml_application}}
\label{tab:ml}
\resizebox{.9\textwidth}{!}{
\begin{tabular}{c|c|c|c|c}
\toprule
Application  & Method  & Role of LLM  & Specific Problems or Tasks \\


\midrule
\multirow{6}{*}{Reinforcement Learning} & LFG~\citep{shah2023navigation}  & LLMaP   & ObjectNav Tasks, Real-world Tasks  \\
\cmidrule{2-4}
& SLINVIT~\citep{zhang2024can}& LLMaP     & ALFWorld Tasks, InterCode Tasks, BlocksWorld Tasks \\
\cmidrule{2-4}
& MEDIC~\citep{bhambri2024efficient}  & LLMaP     & BabyAI Tasks   \\
\cmidrule{2-4}
& Eureka~\citep{ma2024eureka} & LLMaD    & IsaacGym Tasks, Bidexterous Manipulation Tasks \\
\cmidrule{2-4}
& EROM~\citep{narin2024evolutionary}  & LLMaD    & IsaacGym Tasks \\
\cmidrule{2-4}
& LLM-MOE~\citep{du2024mixture}   & LLMaP    & Intelligent Networks   \\

\midrule
\multirow{4}{*}{Neural Architecture Search} & EvoPrompting~\citep{chen2024evoprompting}   & LLMaD    & MNIST Dataset, CLRS Algorithmic Reasoning  \\
\cmidrule{2-4}
& HS-NAS~\citep{jawahar2024llm}   & LLMaP    & Machine Translation Tasks  \\
\cmidrule{2-4}
& LLMatic~\citep{nasir2023llmatic}& LLMaD   & CIFAR-10 Dataset, NAS-bench-201 Benchmarks \\
\cmidrule{2-4}
& LAPT~\citep{zhou2024design} & LLMaD  & NAS201, Trans101, DARTs \\
\cmidrule{2-4}
& LLM-GE~\citep{morris2024llm}& LLMaD    & CIFAR-10 Dataset   \\

\midrule
\multirow{7}{*}{Prompt Tuning}
  & APE \citep{zhou2022large}  & LLMaO   & Instruction Induction, BBH\\
\cmidrule{2-4}
  & OPRO \citep{yang2024large} & LLMaO   & GSM8K, BBH  \\
\cmidrule{2-4}
  & APO \citep{pryzant2023automatic} & LLMaO & NLP Benchmark Classification Tasks  \\
\cmidrule{2-4}
  & GPO \citep{tang2024unleashing} & LLMaO   & Reasoning, Knowledge-intensive, NLP Tasks \\
\cmidrule{2-4}
  & MaaO \citep{guo2024two}& LLMaO   & NLU Tasks, Image Classification Tasks \\
\cmidrule{2-4}
  & EvoPrompt \citep{guo2023connecting} & LLMaO  & Language Understanding and Generation Tasks, BBH  \\
  \cmidrule{2-4}
  & StrategyLLM \citep{gao2023strategyllm} & Mixed   & Reasoning Tasks  \\
  
\midrule
\multirow{5}{*}{Graph Learning} & LLM-Critical~\citep{mao2024identify}& LLMaD    & Critical Node Identification   \\
\cmidrule{2-4}
& LLM-GNN~\citep{chen2023label}   & LLMaP   & Label-free Node Classification \\
\cmidrule{2-4}
& ReStruct~\citep{chen2024hin}& LLMaE    & Meta-structure Discovery   \\
\cmidrule{2-4}
& AutoAlign~\citep{zhou2024autoalign} & LLMaE     & Entity Type Inference  \\
\cmidrule{2-4}
& KSL~\citep{feng2023knowledge}   & LLMaP    & Knowledge Search   \\
\cmidrule{2-4}
& AutoSGNN \citep{mo2025autosgnn} & LLMaD & GNNs \\
\midrule
\multirow{2}{*}{Dataset Labeling}   & Inter-Classier~\citep{chiquier2024evolving} & LLMaO    & iNaturalist Datasets, KikiBouba datasets   \\
\cmidrule{2-4}
& DataSculpt~\citep{guan2023can}  & LLMaP   & Label Function Design  \\
\midrule

\multirow{4}{*}{Other Applications} & LLM2FEA~\citep{wong2024llm2fea}& LLMaP   & Objective-oriented Generation  \\
\cmidrule{2-4}
& tnGPS~\citep{zeng2024tnGPS} & LLMaD    & Tensor Network Structure Search   \\
\cmidrule{2-4}
& L-AutoDA \citep{guo2024autoda}  & LLMaD    & Adversarial Attack   \\
\cmidrule{2-4}
& L-SFE \citep{wang2025llm}  & LLMaD    & Causal Structure Learning \\
\bottomrule
\end{tabular}

}
\vspace{-1em}
\end{table*}

\subsubsection{Other Applications}
Other applications of LLMs extend to a myriad of machine learning tasks. Notably, \citet{wong2024llm2fea} capitalize on multi-task learning and the iterative refinement of prompts to foster innovative design approaches. \citet{zeng2024tnGPS} integrate LLMs with evolutionary search to develop a heuristic function aimed at efficiently sifting through candidate tensor sets representing the primal tensor network. Furthermore, \citet{guo2024autoda} have blazed a trail in employing LLMs to generate novel decision-based adversarial attack algorithms, thus opening up a new diagram for the automatic assessment of model robustness.

\subsection{Scientific and Symbolic Domains}

This subsection is dedicated to exploring LLM-based scientific discoveries. Rather than focusing on classic algorithm design, the works discussed here demonstrate how LLMs function as algorithmic engines to create search strategies, predict performance, and generate symbolic expressions to solve complex scientific problems. Table.~\ref{tab:sd_application} lists the related works in this domain. 

\subsubsection{Symbolic Regression}
In the realm of scientific discovery, LLMs are usually adopted for equation or functioin search. Notably, \citet{du2024llm4ed} introduce LLM4ED, a framework that employs iterative strategies, including a black-box optimizer and evolutionary operators, to generate and optimize equations. This approach has shown significant advancements in stability and usability for uncovering physical laws from nonlinear dynamic systems. Similarly, \citet{shojaee2024llm} present LLM-SR, which combines LLMs' extensive scientific knowledge and code generation capabilities with evolutionary search. This framework excels in proposing and refining initial equation structures based on physical understanding, outperforming traditional symbolic regression methods in discovering physically accurate equations across multiple scientific domains. A bilevel optimization framework, named SGA, is introduce by \citet{ma2024llm}. It merges the abstract reasoning capabilities of LLMs with the computational power of simulations. This integration facilitates hypothesis generation and discrete reasoning with simulations for experimental feedback and optimization of continuous variables, leading to improved performance.

\subsubsection{Chemistry, Biology \& Physics}

In the field of chemistry, LLMs can be applied not only to conventional molecular generation and design \citep{ishida2024large,bhattacharya2024large}, but also to specialized areas such as drug molecule design \citep{ye2024novo}, chemical reaction prediction and optimization \citep{rankovic2023bochemian}, and catalyst design \citep{wang2024catalm}, providing customized solutions. Furthermore, LLMs have also shown promising application prospects in materials discovery~\citep{jia2024llmatdesign,zhang2024honeycomb}, synthesis route planning~\citep{liu2024multimodal}, green chemistry~\citep{ruff2024using}, and other areas. These studies demonstrate the advantages of LLMs in molecular representation and optimization. 

In biology, LLMs are increasingly being used for tasks such as protein engineering \cite{shen2024toursynbio}, enzyme design \cite{teukam2024integrating}, and biological sequence analysis \cite{feng2024large}. By combining LLMs with vast amounts of biological data, they can more accurately predict interactions between biological molecules and significantly improve the efficiency of bioinformatics workflows. This has important implications for drug discovery and therapeutic protein design. The unique sequence generation and optimization capabilities of LLMs offer new possibilities for tackling combinatorial optimization challenges in biomacromolecular design.

Although applications in physics are relatively fewer, some emerging work has demonstrated the broad prospects of LLMs. \citet{pan2024quantum} use multi-step prompt templates to prove that LLMs can perform complex analytical calculations in theoretical physics. Quantum computing algorithm design, physics simulation optimization, and computational methods in condensed matter.



\subsubsection{Mechanics}

MechAgents~\citep{ni2024mechagents} introduces a class of physics-inspired generative machine learning platforms that utilize multiple LLMs to solve mechanics problems, such as elasticity, by autonomously collaborating to write, execute, and self-correct code using finite element methods. Moreover, \citet{du2024large} adopt LLMs in automatically discovering governing equations from data, utilizing the models' generation and reasoning capabilities to iteratively refine candidate equations. This approach, tested on various nonlinear systems including the Burgers, Chafee–Infante, and Navier–Stokes equations, demonstrates a superior ability to uncover correct physical laws and shows better generalization on unseen data compared to other models. AutoTurb~\citep{zhang2024autoturb}, on the other hand, adopts LLMs in an evolutionary framework to automatically design and search for turbulence model in computational fluid dynamics. Furthermore, \citet{buehler2023melm} study LLM-based methods for forward and inverse mechanics problems, including bio-inspired hierarchical honeycomb design, carbon nanotube mechanics, and protein unfolding.

\begin{table*}[htbp]
\centering
\caption{An Overview of Science Discovery Applications Utilizing Language Models Across Various Domains and Tasks.\label{tab:sd_application}}
\label{tab:sd}
\resizebox{0.9\textwidth}{!}{
\begin{tabular}{c|c|c|c}
\toprule
Application  & Method  & Role of LLM & Specific Problems or Tasks \\
\midrule
\multirow{3}{*}{ General Scientific Equation Discovery } 
& Bilevel \citep{ma2024llm}   & LLMaD & Physical Scientific Discovery \\
\cmidrule{2-4}

& LLM-SR \citep{shojaee2024llm} & LLMaD & Scientific Equation Discovery\\
\cmidrule{2-4}

& LLM4ED \citep{du2024llm4ed} & LLMaD & Equation Discovery  \\
\midrule

\multirow{8}{*}{Chemical }  
& ChatChemTS~\citep{ishida2024large} & LLMaD & Molecule   Design \\
\cmidrule{2-4}

& Debjyoti Bhattacharya, et al.~\citep{bhattacharya2024large} & LLMaO & Molecule Design   \\
\cmidrule{2-4}

& Agustinus Kristiadi, et al.~\citep{kristiadi2024sober}   & LLMaE & Molecule Design  \\
\cmidrule{2-4}

& BoChemian~\citep{rankovic2023bochemian} & LLMaE & Chemical Reaction \\
\cmidrule{2-4}

& Multi-modal MoLFormer \citep{soares2024capturing}& LLMaP & Battery Electrolytes Formulation   Design \\
\cmidrule{2-4}

& CataLM \citep{wang2024catalm}& LLMaP & Catalyst Design   \\
\cmidrule{2-4}

& Gavin Ye \citep{ye2024novo}  & LLMaD & Drug Design   \\
\cmidrule{2-4}

& DrugAssist \citep{ye2023drugassist}& LLMaO & Drug Design  \\
\midrule

\multirow{5}{*}{Biology}   
& MLDE~\citep{tran2024protein}  & LLMaP & Protein Design\\
\cmidrule{2-4}

& Prollama~\citep{lv2024prollama} & Mixed & Protein Design \\\cmidrule{2-4}

& X-LoRA \citep{buehler2024x}& LLMaP & Protein Design\\
\cmidrule{2-4}

& CodonBERT \citep{li2023codonbert} & LLMaP & mRNA Design and Optimization  \\
\cmidrule{2-4}

& Revisiting-PLMs \citep{hu2022exploring} & LLMaP & Protein Function Prediction   \\
\midrule

\multirow{4}{*}{Mechanics} 

& MechAgents~\citep{ni2024mechagents}                 & LLMaD        & Mechanics Design   \\
\cmidrule{2-4}

& MeLM~\citep{buehler2023melm}   & LLMaP, LLMaD & Carbon Nanotubes and Proteins Design    \\
\cmidrule{2-4}

& Mengge Du, et al.~\citep{du2024large} & LLMaD        & Nonlinear Dynamics Equation Discovery \\
\cmidrule{2-4}

& AutoTurb~\citep{zhang2024autoturb} & LLMaD & Computational Fliud Dynamics \\
  \bottomrule
\end{tabular}
}
\end{table*}

\section{Challenges and Future Directions}\label{sec6:challenges}
\subsection{Scalability}
A primary limitation of LLMs in algorithm design is their scalability. LLMs operate within a fixed context window, which restricts the amount of information they can process simultaneously. This constraint is particularly problematic for complex algorithmic tasks that involve detailed specifications or large inputs. Furthermore, even with an adequate context window, LLMs often struggle to accurately comprehend and reason over long sequences of information and generate a long solution~\cite{chen2024design}.

While the growing capabilities of LLMs and advancements in reasoning models~\cite{guo2025deepseek} have improved performance on tasks that can be mapped to their prior knowledge, this strength offers little advantage for novel algorithm design problems~\cite{liu2025interpretable}. For instance, when used as optimizers, LLMs tend to perform well only on low-dimensional problems~\citep{yang2024large}. Similarly, in the LLMaD paradigm, LLMs are often limited to generating individual components or heuristic functions rather than a complete algorithm framework~\citep{liu2024evolution}.

To mitigate these scalability challenges, some researchers have proposed hybrid methods. For example, \citet{novikov2025alphaevolve} introduce an approach that combines targeted code modifications using diff blocks with complete code regeneration, allowing for the evolution of more complex programs. Although related research has explored long-code generation~\cite{Ishibashi2024selforganized}, these efforts typically prioritize code correctness and general software engineering over the specific challenge of algorithmic scalability.

\subsection{Generalization}
Another major challenge lies in generalization across different distributions and problem types. Existing methods often exhibit strong performance within their training distribution but fail to generalize to unseen instances~\citep{sim2025beyond}. Designing a single heuristic that performs optimally across diverse instance distributions is inherently difficult~\citep{sim2025beyond,shi2025generalizable}. For example, \citet{sim2025beyond} reveal that algorithms designed on synthetic bin packing instances perform poorly on bin packing instances with real-world distributions. Similarly, \citet{shi2025generalizable} show that an algorithm designed for TSP instances with a fixed size can hardly generalize to TSP instances with other sizes.

Several approaches aim to enhance cross-distribution generalization. \citet{shi2025generalizable} adopt a meta-learning framework that promotes adaptation across distributions. \citet{zhang2025llm} partition the overall problem class into subclasses based on instance features, enabling differentiated and automated heuristic design for each subclass. In contrast to designing a single optimal algorithm, EoH-S~\citep{liu2025eoh} proposes the concept of heuristic set design, which creates a portfolio of complementary algorithms to improve overall robustness across heterogeneous distributions.

Beyond distribution generalization, cross-problem generalization presents an even greater challenge, particularly for LLM-driven algorithm design. The majority of existing works focus on developing algorithms for a specific target task, and the resulting algorithm can rarely be applied to solve other problems~\citep{liu2024evolution}. On the one hand, the algorithmic code implementation can not be used for solving other problems. On the other hand, the designed algorithm idea is either too high-level or problem-specific and thus lack cross-problem generalization. This contrasts with commonly used algorithmic frameworks, such as metaheuristic pipelines, which are often applicable to various problems despite differences in specific code implementations, as the core algorithmic idea remains the same~\citep{gendreau2010handbook}. While there has been some discussion on designing general metaheuristics~\citep{pluhacek2023leveraging}, this work has remained at a high level without comprehensive evaluation. Consequently, cross-problem algorithm design is a promising and challenging direction for future research.

\subsection{Interpretability}
Due to the black-box nature of LLMs, it is often difficult to trace how specific inputs influence the generated outputs. Moreover, we lack a deep understanding of the iterative algorithm design process itself~\citep{allen2023physics,liu2025large}.

Recent studies have begun to address this challenge by analyzing the algorithmic landscape of LLM-driven design. For instance, \citet{liu2025large} employ a simpler, interpretable model to approximate the behavior of LLMs when used as optimizers for multi-objective problems. To visualize the design process, \citet{van2025code} propose Code Evolution Graphs (CEGs), which map the evolution of code using Abstract Syntax Tree (AST) features. \citet{liu2025fitness} introduce a graph-based representation where nodes signify algorithms and edges denote their evolutionary relationships, enabling a structural analysis of the design space. A related and crucial research direction involves developing appropriate similarity metrics to quantify the relationships between generated algorithms, which remains an active area of investigation~\citep{yao2024multi,liu2025fitness}.

Other work has focused on establishing a theoretical basis for these methods. \citet{lee2025convergence} propose to interpret the LLMaO procedure as a finite-state Markov chain, providing a potential avenue for formal analysis.

\subsection{Efficiency}
Efficiency is another major challenge in applying LLMs to algorithm design, encompassing both efficient iterative algorithm search and evaluation reduction.

While early methods relied on simple evolutionary search~\cite{liu2024evolution}, more sophisticated strategies have been introduced to improve search efficiency. These include reflective reasoning~\cite{ye2024reevo}, Monte Carlo Tree Search~\cite{zhengmonte}, and ensemble methods~\cite{novikov2025alphaevolve}. Recent attempts have also explored fine-tuning LLMs specifically for algorithm design, using both offline~\cite{liu2025fine} and online~\cite{surina2025algorithm,huang2025calm} approaches. For instance, \citet{liu2025fine} constructed diverse algorithm pairs and employed Direct Preference Optimization (DPO) to offline fine-tune a smaller model. This specialized model demonstrated higher inference speeds and achieved results competitive with much larger models. In an online setting, \citet{huang2025calm} utilized reinforcement learning to dynamically update the model during the algorithm search, leading to faster convergence and superior final algorithms within a fixed computational budget. However, these fine-tuned models are typically specialized for code generation on a single task. A key open question is how to accelerate the entire algorithm design process in a way that generalizes across diverse problem domains.

Furthermore, real-world algorithm design often requires computationally expensive evaluations. This expense arises from two factors: the large number of test instances required for robust assessment and the time-consuming nature of each individual evaluation, which may involve complex simulations or searches. Recent work has attempted to reduce this evaluation cost by enabling the LLM to predict an algorithm's quality and terminate unpromising evaluations early~\cite{jawahar2024llm}. Despite these efforts, there remains a need for principled approaches that can effectively balance the trade-off between exploration efficiency and performance reliability.

\subsection{Benchmarking}
Benchmarking is essential for ensuring fairness, reproducibility, and standardization across studies, facilitating both qualitative and quantitative comparisons. Despite rapid progress in LLM4AD, the field currently lacks standardized benchmarks for either algorithm design tasks or evaluation pipelines assessing LLM capabilities in this context.

While related benchmarks have been developed for mathematical reasoning~\citep{lin2024atg}, planning~\citep{valmeekam2024planbench}, and classical algorithmic tasks~\citep{velivckovic2022clrs}, they do not fully capture the nuances of LLM-driven design. Recently, efforts have begun to develop benchmarks specifically for LLM-driven algorithm design. For example, \citet{sun2025co} and \citet{chen2025heurigym} introduce benchmark suites focused on combinatorial optimization. \citet{van2025blade} developed a benchmark for designing metaheuristics for black-box optimization. Broadening the scope, \citet{liu2024llm4ad} proposed a collection of tasks spanning diverse domains, from optimization and machine learning to scientific discovery.

Despite this progress, establishing rigorous and widely accepted benchmarks with standard settings remains a significant challenge. The development of datasets, unified evaluation protocols, and transparent reporting practices is crucial. Such standards will not only enhance reproducibility but also catalyze innovation in this emerging research area.

\section{Conclusion}\label{sec7:conclusion}

In this survey, we have provided a systematic review of the emerging field of algorithm design with large language models (LLM4AD). We categorized existing works into four paradigms based on the LLM's primary role: LLM as Optimizer (LLMaO), LLM as Predictor (LLMaP), LLM as Extractor (LLMaE), and LLM as Designer (LLMaD). The LLMaO paradigm employs the LLM as a black-box optimizer to generate solutions, while LLMaP uses it as a surrogate model for prediction. LLMaE leverages the LLM to derive semantic features from unstructured data to inform an algorithm, and LLMaD directly generates algorithmic components or designs the entire algorithm. Furthermore, we have mapped works to the three core stages of the algorithm design, i.e., ideation, implementation, and evaluation, highlighting both progress and limitations. Finally, we summarized key application domains and identified critical open challenges, including scalability, generalization, interpretability, efficiency, and benchmarking, to help guide future research in this area.












\bibliographystyle{ACM-Reference-Format}
\bibliography{manuscript}

@book{kleinberg2006algorithm,
  title={Algorithm Design},
  author={Kleinberg, J},
  volume={92},
  year={2006},
  publisher={Pearson Education}
}

@book{cormen2022introduction,
  title={Introduction to algorithms},
  author={Cormen, Thomas H and Leiserson, Charles E and Rivest, Ronald L and Stein, Clifford},
  year={2022},
  publisher={MIT press}
}

@inproceedings{gurkan2025lear,
  title={LEAR: LLM-Driven Evolution of Agent-Based Rules},
  author={Gurkan, Can and Jwalapuram, Narasimha Karthik and Wang, Kevin and Danda, Rudy and Rasmussen, Leif and Chen, John and Wilensky, Uri},
  booktitle={Proceedings of the Genetic and Evolutionary Computation Conference Companion},
  pages={2309--2326},
  year={2025}
}

@article{buehler2023melm,
  title={MeLM, a generative pretrained language modeling framework that solves forward and inverse mechanics problems},
  author={Buehler, Markus J},
  journal={Journal of the Mechanics and Physics of Solids},
  volume={181},
  pages={105454},
  year={2023},
  publisher={Elsevier}
}

@article{yu2024deep,
  title={Deep insights into automated optimization with large language models and evolutionary algorithms},
  author={Yu, He and Liu, Jing},
  journal={arXiv preprint arXiv:2410.20848},
  year={2024}
}

@article{jiang2024survey,
  title={A Survey on Large Language Models for Code Generation},
  author={Jiang, Juyong and Wang, Fan and Shen, Jiasi and Kim, Sungju and Kim, Sunghun},
  journal={arXiv preprint arXiv:2406.00515},
  year={2024}
}

@inproceedings{liu2024evolution,
	author    = {Liu, Fei and Xialiang, Tong and Yuan, Mingxuan and Lin, Xi and Luo, Fu and Wang, Zhenkun and Lu, Zhichao and Zhang, Qingfu},
	booktitle = {Forty-first International Conference on Machine Learning},
	title     = {Evolution of Heuristics: Towards Efficient Automatic Algorithm Design Using Large Language Model},
	year      = {2024}
}

@article{chen2024evoprompting,
	author  = {Chen, Angelica and Dohan, David and So, David},
	journal = {Advances in Neural Information Processing Systems},
	title   = {EvoPrompting: language models for code-level neural architecture search},
	volume  = {36},
	year    = {2024}
}

@inproceedings{velivckovic2022clrs,
	author       = {Veli{\v{c}}kovi{\'c}, Petar and Badia, Adri{\`a} Puigdom{\`e}nech and Budden, David and Pascanu, Razvan and Banino, Andrea and Dashevskiy, Misha and Hadsell, Raia and Blundell, Charles},
	booktitle    = {International Conference on Machine Learning},
	organization = {PMLR},
	pages        = {22084--22102},
	title        = {The CLRS algorithmic reasoning benchmark},
	year         = {2022}
}

@article{valmeekam2024planbench,
	author  = {Valmeekam, Karthik and Marquez, Matthew and Olmo, Alberto and Sreedharan, Sarath and Kambhampati, Subbarao},
	journal = {Advances in Neural Information Processing Systems},
	title   = {Planbench: An extensible benchmark for evaluating large language models on planning and reasoning about change},
	volume  = {36},
	year    = {2024}
}

@inproceedings{lin2024atg,
  title={ATG: Benchmarking Automated Theorem Generation for Generative Language Models},
  author={Lin, Xiaohan and Cao, Qingxing and Huang, Yinya and Yang, Zhicheng and Liu, Zhengying and Li, Zhenguo and Liang, Xiaodan},
  booktitle={Findings of the Association for Computational Linguistics: NAACL 2024},
  pages={4465--4480},
  year={2024}
}

@inproceedings{
guo2023connecting,
title={Connecting Large Language Models with Evolutionary Algorithms Yields Powerful Prompt Optimizers},
author={Qingyan Guo and Rui Wang and Junliang Guo and Bei Li and Kaitao Song and Xu Tan and Guoqing Liu and Jiang Bian and Yujiu Yang},
booktitle={The Twelfth International Conference on Learning Representations},
year={2024}
}

@article{ma2024large,
	author  = {Ma, Ruotian and Wang, Xiaolei and Zhou, Xin and Li, Jian and Du, Nan and Gui, Tao and Zhang, Qi and Huang, Xuanjing},
	journal = {arXiv preprint arXiv:2402.02101},
	title   = {Are Large Language Models Good Prompt Optimizers?},
	year    = {2024}
}

@inproceedings{zhou2022large,
  title={Large language models are human-level prompt engineers},
  author={Zhou, Yongchao and Muresanu, Andrei Ioan and Han, Ziwen and Paster, Keiran and Pitis, Silviu and Chan, Harris and Ba, Jimmy},
  booktitle={The eleventh international conference on learning representations},
  year={2022}
}

@article{liu2023algorithm,
	author  = {Liu, Fei and Tong, Xialiang and Yuan, Mingxuan and Zhang, Qingfu},
	journal = {arXiv preprint arXiv:2311.15249},
	title   = {Algorithm evolution using large language model},
	year    = {2023}
}

@inproceedings{zhang2024understanding,
  title={Understanding the Importance of Evolutionary Search in Automated Heuristic Design with Large Language Models},
  author={Zhang, Rui and Liu, Fei and Lin, Xi and Wang, Zhenkun and Lu, Zhichao and Zhang, Qingfu},
  booktitle={International Conference on Parallel Problem Solving from Nature},
  pages={185--202},
  year={2024},
  organization={Springer}
}

@inproceedings{
ma2024llm,
title={{LLM} and Simulation as Bilevel Optimizers: A New Paradigm to Advance Physical Scientific Discovery},
author={Pingchuan Ma and Tsun-Hsuan Wang and Minghao Guo and Zhiqing Sun and Joshua B. Tenenbaum and Daniela Rus and Chuang Gan and Wojciech Matusik},
booktitle={Forty-first International Conference on Machine Learning},
year={2024}
}

@article{lv2024prollama,
  title={Prollama: A protein large language model for multi-task protein language processing},
  author={Lv, Liuzhenghao and Lin, Zongying and Li, Hao and Liu, Yuyang and Cui, Jiaxi and Chen, Calvin Yu-Chian and Yuan, Li and Tian, Yonghong},
  journal={IEEE Transactions on Artificial Intelligence},
  year={2025},
  publisher={IEEE}
}

@article{teukam2024integrating,
  title={Integrating genetic algorithms and language models for enhanced enzyme design},
  author={Nana Teukam, Yves Gaetan and Zipoli, Federico and Laino, Teodoro and Criscuolo, Emanuele and Grisoni, Francesca and Manica, Matteo},
  journal={Briefings in bioinformatics},
  volume={26},
  number={1},
  pages={bbae675},
  year={2025},
  publisher={Oxford University Press}
}

@inproceedings{shen2024toursynbio,
  title={Toursynbio: A multi-modal large model and agent framework to bridge text and protein sequences for protein engineering},
  author={Shen, Yiqing and Chen, Zan and Mamalakis, Michail and Liu, Yungeng and Li, Tianbin and Su, Yanzhou and He, Junjun and Li{\`o}, Pietro and Wang, Yu Guang},
  booktitle={2024 IEEE International Conference on Bioinformatics and Biomedicine (BIBM)},
  pages={2382--2389},
  year={2024},
  organization={IEEE}
}

@article{zhang2024autoturb,
  title={AutoTurb: Using large language models for automatic algebraic turbulence model discovery},
  author={Zhang, Yu and Zheng, Kefeng and Liu, Fei and Zhang, Qingfu and Wang, Zhenkun},
  journal={Physics of Fluids},
  volume={37},
  number={1},
  year={2025},
  publisher={AIP Publishing}
}

@article{jia2024llmatdesign,
  title={LLMatDesign: Autonomous Materials Discovery with Large Language Models},
  author={Jia, Shuyi and Zhang, Chao and Fung, Victor},
  journal={arXiv preprint arXiv:2406.13163},
  year={2024}
}

@inproceedings{zhang2024honeycomb,
  title={HoneyComb: A Flexible LLM-Based Agent System for Materials Science},
  author={Zhang, Huan and Song, Yu and Hou, Ziyu and Miret, Santiago and Liu, Bang},
  booktitle={Findings of the Association for Computational Linguistics: EMNLP 2024},
  pages={3369--3382},
  year={2024}
}

@inproceedings{
liu2024multimodal,
title={Multimodal Large Language Models for Inverse Molecular Design with Retrosynthetic Planning},
author={Gang Liu and Michael Sun and Wojciech Matusik and Meng Jiang and Jie Chen},
booktitle={The Thirteenth International Conference on Learning Representations},
year={2025}
}

@article{ruff2024using,
  title={Using ChatGPT for Method Development and Green Chemistry Education in Upper-Level Laboratory Courses},
  author={Ruff, Emily F and Franz, Jeanne L and West, Joseph K},
  journal={Journal of Chemical Education},
  volume={101},
  number={8},
  pages={3224--3232},
  year={2024},
  publisher={ACS Publications}
}

@article{feng2024large,
  title={Large language models for biomolecular analysis: From methods to applications},
  author={Feng, Ruijun and Zhang, Chi and Zhang, Yang},
  journal={TrAC Trends in Analytical Chemistry},
  pages={117540},
  year={2024},
  publisher={Elsevier}
}

@book{gendreau2010handbook,
	author    = {Gendreau, Michel and Potvin, Jean-Yves and others},
	publisher = {Springer},
	title     = {Handbook of metaheuristics},
	volume    = {2},
	year      = {2010}
}

@article{wu2024evolutionary,
  title={Evolutionary computation in the era of large language model: Survey and roadmap},
  author={Wu, Xingyu and Wu, Sheng-hao and Wu, Jibin and Feng, Liang and Tan, Kay Chen},
  journal={IEEE Transactions on Evolutionary Computation},
  year={2024},
  publisher={IEEE}
}

@inproceedings{li2024auto,
	author    = {Li, Hao and Yang, Xue and Wang, Zhaokai and Zhu, Xizhou and Zhou, Jie and Qiao, Yu and Wang, Xiaogang and Li, Hongsheng and Lu, Lewei and Dai, Jifeng},
	booktitle = {Proceedings of the IEEE/CVF Conference on Computer Vision and Pattern Recognition},
	pages     = {16426--16435},
	title     = {Auto mc-reward: Automated dense reward design with large language models for minecraft},
	year      = {2024}
}

@article{van2024llamea,
  title={Llamea: A large language model evolutionary algorithm for automatically generating metaheuristics},
  author={van Stein, Niki and B{\"a}ck, Thomas},
  journal={IEEE Transactions on Evolutionary Computation},
  year={2024},
  publisher={IEEE}
}

@article{van2024loop,
  title={In-the-loop hyper-parameter optimization for llm-based automated design of heuristics},
  author={van Stein, Niki and Vermetten, Diederick and B{\"a}ck, Thomas},
  journal={ACM Transactions on Evolutionary Learning},
  year={2024},
  publisher={ACM New York, NY}
}

@inproceedings{
hu2024automated,
title={Automated Design of Agentic Systems},
author={Shengran Hu and Cong Lu and Jeff Clune},
booktitle={The Thirteenth International Conference on Learning Representations},
year={2025}
}

@inproceedings{guo2024autoda,
  title={L-autoda: Large language models for automatically evolving decision-based adversarial attacks},
  author={Guo, Ping and Liu, Fei and Lin, Xi and Zhao, Qingchuan and Zhang, Qingfu},
  booktitle={Proceedings of the Genetic and Evolutionary Computation Conference Companion},
  pages={1846--1854},
  year={2024}
}

@inproceedings{yao2024evolve,
  title={Evolve cost-aware acquisition functions using large language models},
  author={Yao, Yiming and Liu, Fei and Cheng, Ji and Zhang, Qingfu},
  booktitle={International Conference on Parallel Problem Solving from Nature},
  pages={374--390},
  year={2024},
  organization={Springer}
}

@inproceedings{morris2024llm,
  title={Llm guided evolution-the automation of models advancing models},
  author={Morris, Clint and Jurado, Michael and Zutty, Jason},
  booktitle={Proceedings of the Genetic and Evolutionary Computation Conference},
  pages={377--384},
  year={2024}
}

@article{brahmachary2024large,
  title={Large language model-based evolutionary optimizer: Reasoning with elitism},
  author={Brahmachary, Shuvayan and Joshi, Subodh M and Panda, Aniruddha and Koneripalli, Kaushik and Sagotra, Arun Kumar and Patel, Harshil and Sharma, Ankush and Jagtap, Ameya D and Kalyanaraman, Kaushic},
  journal={Neurocomputing},
  volume={622},
  pages={129272},
  year={2025},
  publisher={Elsevier}
}

@inproceedings{
li2023codonbert,
title={Codon{BERT}: Large Language Models for m{RNA} design and optimization},
author={Sizhen Li and Saeed Moayedpour and Ruijiang Li and Michael Bailey and Saleh Riahi and Milad Miladi and Jacob Miner and Dinghai Zheng and Jun Wang and Akshay Balsubramani and Khang Tran and Minnie and Monica Wu and Xiaobo Gu and Ryan Clinton and Carla Asquith and Joseph Skaleski and Lianne Boeglin and Sudha Chivukula and Anusha Dias and Fernando Ulloa Montoya and Vikram Agarwal and Ziv Bar-Joseph and Sven Jager},
booktitle={NeurIPS 2023 Generative AI and Biology (GenBio) Workshop},
year={2023}
}

@article{egami2024using,
	author  = {Egami, Naoki and Hinck, Musashi and Stewart, Brandon and Wei, Hanying},
	journal = {Advances in Neural Information Processing Systems},
	title   = {Using imperfect surrogates for downstream inference: Design-based supervised learning for social science applications of large language models},
	volume  = {36},
	year    = {2024}
}

@inproceedings{wang2024resllm,
  title={Resllm: Large language models are strong resource selectors for federated search},
  author={Wang, Shuai and Zhuang, Shengyao and Koopman, Bevan and Zuccon, Guido},
  booktitle={Companion Proceedings of the ACM on Web Conference 2025},
  pages={1360--1364},
  year={2025}
}

@inproceedings{shah2023navigation,
	author       = {Shah, Dhruv and Equi, Michael Robert and Osi{\'n}ski, B{\l}a{\.z}ej and Xia, Fei and Ichter, Brian and Levine, Sergey},
	booktitle    = {Conference on Robot Learning},
	organization = {PMLR},
	pages        = {2683--2699},
	title        = {Navigation with large language models: Semantic guesswork as a heuristic for planning},
	year         = {2023}
}

@inproceedings{lange2024large,
	author    = {Lange, Robert and Tian, Yingtao and Tang, Yujin},
	booktitle = {Proceedings of the Genetic and Evolutionary Computation Conference Companion},
	pages     = {579--582},
	title     = {Large language models as evolution strategies},
	year      = {2024}
}

@article{nie2024importance,
  title={The importance of directional feedback for llm-based optimizers},
  author={Nie, Allen and Cheng, Ching-An and Kolobov, Andrey and Swaminathan, Adith},
  journal={arXiv preprint arXiv:2405.16434},
  year={2024}
}

@article{duan2025ealg,
  title={EALG: Evolutionary Adversarial Generation of Language Model-Guided Generators for Combinatorial Optimization},
  author={Duan, Ruibo and Liu, Yuxin and Dong, Xinyao and Fan, Chenglin},
  journal={arXiv preprint arXiv:2506.02594},
  year={2025}
}

@inproceedings{d2024exploring,
  title={Exploring LLM-driven explanations for quantum algorithms},
  author={d'Aloisio, Giordano and Fortz, Sophie and Hanna, Carol and Fortunato, Daniel and Bensoussan, Avner and Mendiluze Usandizaga, E{\~n}aut and Sarro, Federica},
  booktitle={Proceedings of the 18th ACM/IEEE International Symposium on Empirical Software Engineering and Measurement},
  pages={475--481},
  year={2024}
}

@article{li2025cocoevo,
  title={CoCoEvo: Co-Evolution of Programs and Test Cases to Enhance Code Generation},
  author={Li, Kefan and Yuan, Yuan and Yu, Hongyue and Guo, Tingyu and Cao, Shijie},
  journal={IEEE Transactions on Evolutionary Computation},
  year={2025},
  publisher={IEEE}
}

@article{jiang2024bridging,
  title={Bridging Large Language Models and Optimization: A Unified Framework for Text-attributed Combinatorial Optimization},
  author={Jiang, Xia and Wu, Yaoxin and Wang, Yuan and Zhang, Yingqian},
  journal={arXiv preprint arXiv:2408.12214},
  year={2024}
}

@inproceedings{
yang2024large,
title={Large Language Models as Optimizers},
author={Chengrun Yang and Xuezhi Wang and Yifeng Lu and Hanxiao Liu and Quoc V Le and Denny Zhou and Xinyun Chen},
booktitle={The Twelfth International Conference on Learning Representations},
year={2024},
}

@article{hu2025discovering,
  title={Discovering Interpretable Programmatic Policies via Multimodal LLM-assisted Evolutionary Search},
  author={Hu, Qinglong and Tong, Xialiang and Yuan, Mingxuan and Liu, Fei and Lu, Zhichao and Zhang, Qingfu},
  journal={arXiv preprint arXiv:2508.05433},
  year={2025}
}

@article{jin2018data,
  title={Data-driven evolutionary optimization: An overview and case studies},
  author={Jin, Yaochu and Wang, Handing and Chugh, Tinkle and Guo, Dan and Miettinen, Kaisa},
  journal={IEEE Transactions on Evolutionary Computation},
  volume={23},
  number={3},
  pages={442--458},
  year={2018},
  publisher={IEEE}
}

@article{laporte1992traveling,
  title={The traveling salesman problem: An overview of exact and approximate algorithms},
  author={Laporte, Gilbert},
  journal={European Journal of Operational Research},
  volume={59},
  number={2},
  pages={231--247},
  year={1992},
  publisher={Elsevier}
}

@inproceedings{
chen2023label,
title={Label-free Node Classification on Graphs with Large Language Models ({LLM}s)},
author={Zhikai Chen and Haitao Mao and Hongzhi Wen and Haoyu Han and Wei Jin and Haiyang Zhang and Hui Liu and Jiliang Tang},
booktitle={The Twelfth International Conference on Learning Representations},
year={2024}
}

@article{wong2024generative,
	author  = {Wong, Melvin and Rios, Thiago and Menzel, Stefan and Ong, Yew Soon},
	journal = {arXiv preprint arXiv:2406.09143},
	title   = {Generative AI-based Prompt Evolution Engineering Design Optimization With Vision-Language Model},
	year    = {2024}
}

@inproceedings{bhambri2024efficient,
  title={Efficient reinforcement learning via large language model-based search},
  author={Bhambri, Siddhant and Bhattacharjee, Amrita and Kambhampati, Subbarao and others},
  booktitle={NeurIPS 2024 Workshop on Open-World Agents},
  year={2024}
}

@inproceedings{zhang2024direct,
  title={Direct Preference Optimization of Video Large Multimodal Models from Language Model Reward},
  author={Zhang, Ruohong and Gui, Liangke and Sun, Zhiqing and Feng, Yihao and Xu, Keyang and Zhang, Yuanhan and Fu, Di and Li, Chunyuan and Hauptmann, Alexander G and Bisk, Yonatan and others},
  booktitle={Proceedings of the 2025 Conference of the Nations of the Americas Chapter of the Association for Computational Linguistics: Human Language Technologies (Volume 1: Long Papers)},
  pages={694--717},
  year={2025}
}

@article{alipour2024data,
	author  = {Alipour-Vaezi, Mohammad and Tsui, Kwok-Leung},
	journal = {arXiv preprint arXiv:2404.07434},
	title   = {Data-Driven Portfolio Management for Motion Pictures Industry: A New Data-Driven Optimization Methodology Using a Large Language Model as the Expert},
	year    = {2024}
}

@inproceedings{soares2024capturing,
  title={Capturing formulation design of battery electrolytes with chemical large language model},
  author={Soares, Eduardo and Sharma, Vidushi and Brazil, Emilio Vital and Cerqueira, Renato and Na, Young-Hye},
  booktitle={AI for Accelerated Materials Design-NeurIPS 2023 Workshop},
  year={2023}
}

@article{zhang2024can,
	author  = {Zhang, Shenao and Zheng, Sirui and Ke, Shuqi and Liu, Zhihan and Jin, Wanxin and Yuan, Jianbo and Yang, Yingxiang and Yang, Hongxia and Wang, Zhaoran},
	journal = {arXiv preprint arXiv:2402.16181},
	title   = {How Can LLM Guide RL? A Value-Based Approach},
	year    = {2024}
}

@inproceedings{chacon2024large,
	author    = {Chac{\'o}n Sartori, Camilo and Blum, Christian and Ochoa, Gabriela},
	booktitle = {Proceedings of the Genetic and Evolutionary Computation Conference},
	pages     = {160--168},
	title     = {Large Language Models for the Automated Analysis of Optimization Algorithms},
	year      = {2024}
}

@article{xu2025evospeak,
  title={EvoSpeak: Large Language Models for Interpretable Genetic Programming-Evolved Heuristics},
  author={Xu, Meng and Liu, Jiao and Ong, Yew Soon},
  journal={arXiv preprint arXiv:2510.02686},
  year={2025}
}

@inproceedings{wu2024large,
  title={Large Language Model-Enhanced Algorithm Selection: Towards Comprehensive Algorithm Representation},
  author={Wu, Xingyu and Zhong, Yan and Wu, Jibin and Jiang, Bingbing and Tan, Kay Chen},
  booktitle={Proceedings of the 33rd International Joint Conference on Artificial Intelligence, IJCAI 2024},
  pages={5235--5244},
  year={2024}
}

@article{zhang2025darwin,
  title={Darwin Godel Machine: Open-Ended Evolution of Self-Improving Agents},
  author={Zhang, Jenny and Hu, Shengran and Lu, Cong and Lange, Robert and Clune, Jeff},
  journal={arXiv preprint arXiv:2505.22954},
  year={2025}
}

@inproceedings{mo2025autosgnn,
  title={AutoSGNN: automatic propagation mechanism discovery for spectral graph neural networks},
  author={Mo, Shibing and Wu, Kai and Gao, Qixuan and Teng, Xiangyi and Liu, Jing},
  booktitle={Proceedings of the AAAI Conference on Artificial Intelligence},
  volume={39},
  number={18},
  pages={19493--19502},
  year={2025}
}

@article{ge2025mora,
  title={MORA-LLM: Enhancing Multi-Objective Optimization Recommendation Algorithm by Integrating Large Language Models},
  author={Ge, Yuanyuan and Wu, Likang and Yang, Haipeng and Cheng, Fan and Zhao, Hongke and Zhang, Lei},
  journal={IEEE Transactions on Evolutionary Computation},
  year={2025},
  publisher={IEEE}
}

@article{hao2024large,
  title={Large language models as surrogate models in evolutionary algorithms: A preliminary study},
  author={Hao, Hao and Zhang, Xiaoqun and Zhou, Aimin},
  journal={Swarm and Evolutionary Computation},
  volume={91},
  pages={101741},
  year={2024},
  publisher={Elsevier}
}

@article{xie2025large,
  title={Large language model-driven surrogate-assisted evolutionary algorithm for expensive optimization},
  author={Xie, Lindong and Li, Genghui and Wang, Zhenkun and Chung, Edward and Gong, Maoguo},
  journal={arXiv preprint arXiv:2507.02892},
  year={2025}
}

@article{ahmed2024lm4opt,
  title={LM4OPT: Unveiling the potential of Large Language Models in formulating mathematical optimization problems},
  author={Ahmed, Tasnim and Choudhury, Salimur},
  journal={INFOR: Information Systems and Operational Research},
  pages={1--14},
  year={2024},
  publisher={Taylor \& Francis}
}

@inproceedings{ye2024reevo,
  title={ReEvo: large language models as hyper-heuristics with reflective evolution},
  author={Ye, Haoran and Wang, Jiarui and Cao, Zhiguang and Berto, Federico and Hua, Chuanbo and Kim, Haeyeon and Park, Jinkyoo and Song, Guojie},
  booktitle={Proceedings of the 38th International Conference on Neural Information Processing Systems},
  pages={43571--43608},
  year={2024}
}

@article{yang2025large,
  title={Large Language Model-assisted Meta-optimizer for Automated Design of Constrained Evolutionary Algorithm},
  author={Yang, Xu and Wang, Rui and Li, Kaiwen and Li, Wenhua and Huang, Weixiong},
  journal={arXiv preprint arXiv:2509.13251},
  year={2025}
}

@article{chen2024design,
  title={On the design and analysis of llm-based algorithms},
  author={Chen, Yanxi and Li, Yaliang and Ding, Bolin and Zhou, Jingren},
  journal={arXiv preprint arXiv:2407.14788},
  year={2024}
}

@article{chen2025llm4cmo,
  title={LLM4CMO: Large Language Model-aided Algorithm Design for Constrained Multiobjective Optimization},
  author={Chen, Zhen-Song and Ding, Hong-Wei and Wang, Xian-Jia and Pedrycz, Witold},
  journal={arXiv preprint arXiv:2508.11871},
  year={2025}
}

@article{surina2025algorithm,
  title={Algorithm discovery with llms: Evolutionary search meets reinforcement learning},
  author={Surina, Anja and Mansouri, Amin and Quaedvlieg, Lars and Seddas, Amal and Viazovska, Maryna and Abbe, Emmanuel and Gulcehre, Caglar},
  journal={arXiv preprint arXiv:2504.05108},
  year={2025}
}

@article{huang2025calm,
  title={Calm: Co-evolution of algorithms and language model for automatic heuristic design},
  author={Huang, Ziyao and Wu, Weiwei and Wu, Kui and Wang, Jianping and Lee, Wei-Bin},
  journal={arXiv preprint arXiv:2505.12285},
  year={2025}
}

@article{liu2025fine,
  title={Fine-tuning large language model for automated algorithm design},
  author={Liu, Fei and Zhang, Rui and Lin, Xi and Lu, Zhichao and Zhang, Qingfu},
  journal={arXiv preprint arXiv:2507.10614},
  year={2025}
}

@article{yazdani2025evocut,
  title={EvoCut: Strengthening Integer Programs via Evolution-Guided Language Models},
  author={Yazdani, Milad and Mostajabdaveh, Mahdi and Aref, Samin and Zhou, Zirui},
  journal={arXiv preprint arXiv:2508.11850},
  year={2025}
}

@article{li2025strcmp,
  title={STRCMP: Integrating Graph Structural Priors with Language Models for Combinatorial Optimization},
  author={Li, Xijun and Yang, Jiexiang and Wang, Jinghao and Peng, Bo and Yao, Jianguo and Guan, Haibing},
  journal={arXiv preprint arXiv:2506.11057},
  year={2025}
}

@article{wan2025surgery,
  title={Surgery scheduling based on large language models},
  author={Wan, Fang and Wang, Tao and Wang, Kezhi and Si, Yuanhang and Fondrevelle, Julien and Du, Shuimiao and Duclos, Antoine},
  journal={Artificial Intelligence in Medicine},
  pages={103151},
  year={2025},
  publisher={Elsevier}
}

@article{shinohara2025large,
  title={Large language models as particle swarm optimizers},
  author={Shinohara, Yamato and Xu, Jinglue and Li, Tianshui and Iba, Hitoshi},
  journal={arXiv preprint arXiv:2504.09247},
  year={2025}
}

@article{romera2024mathematical,
	author    = {Romera-Paredes, Bernardino and Barekatain, Mohammadamin and Novikov, Alexander and Balog, Matej and Kumar, M Pawan and Dupont, Emilien and Ruiz, Francisco JR and Ellenberg, Jordan S and Wang, Pengming and Fawzi, Omar and others},
	journal   = {Nature},
	number    = {7995},
	pages     = {468--475},
	publisher = {Nature Publishing Group UK London},
	title     = {Mathematical discoveries from program search with large language models},
	volume    = {625},
	year      = {2024}
}

@article{hemberg2024evolving,
  title={Evolving code with a large language model},
  author={Hemberg, Erik and Moskal, Stephen and O’Reilly, Una-May},
  journal={Genetic Programming and Evolvable Machines},
  volume={25},
  number={2},
  pages={21},
  year={2024},
  publisher={Springer}
}

@inproceedings{
ma2024eureka,
title={Eureka: Human-Level Reward Design via Coding Large Language Models},
author={Yecheng Jason Ma and William Liang and Guanzhi Wang and De-An Huang and Osbert Bastani and Dinesh Jayaraman and Yuke Zhu and Linxi Fan and Anima Anandkumar},
booktitle={The Twelfth International Conference on Learning Representations},
year={2024}
}

@article{bengio2021machine,
	author    = {Bengio, Yoshua and Lodi, Andrea and Prouvost, Antoine},
	journal   = {European Journal of Operational Research},
	number    = {2},
	pages     = {405--421},
	publisher = {Elsevier},
	title     = {Machine learning for combinatorial optimization: a methodological tour d’horizon},
	volume    = {290},
	year      = {2021}
}

@article{wong2024llm2fea,
	author  = {Wong, Melvin and Liu, Jiao and Rios, Thiago and Menzel, Stefan and Ong, Yew Soon},
	journal = {arXiv preprint arXiv:2406.14917},
	title   = {LLM2FEA: Discover Novel Designs with Generative Evolutionary Multitasking},
	year    = {2024}
}

@article{guo2024two,
  title={Two Optimizers Are Better Than One: LLM Catalyst for Enhancing Gradient-Based Optimization},
  author={Guo, Zixian and Liu, Ming and Ji, Zhilong and Bai, Jinfeng and Guo, Yiwen and Zuo, Wangmeng},
  journal={arXiv preprint arXiv:2405.19732},
  year={2024}
}

@inproceedings{tang2024unleashing,
  title={Unleashing the potential of large language models as prompt optimizers: Analogical analysis with gradient-based model optimizers},
  author={Tang, Xinyu and Wang, Xiaolei and Zhao, Wayne Xin and Lu, Siyuan and Li, Yaliang and Wen, Ji-Rong},
  booktitle={Proceedings of the AAAI Conference on Artificial Intelligence},
  volume={39},
  number={24},
  pages={25264--25272},
  year={2025}
}

@inproceedings{
wang2023promptagent,
title={PromptAgent: Strategic Planning with Language Models Enables Expert-level Prompt Optimization},
author={Xinyuan Wang and Chenxi Li and Zhen Wang and Fan Bai and Haotian Luo and Jiayou Zhang and Nebojsa Jojic and Eric Xing and Zhiting Hu},
booktitle={The Twelfth International Conference on Learning Representations},
year={2024}
}

@inproceedings{
pryzant2023automatic,
title={Automatic Prompt Optimization with ''Gradient Descent'' and Beam Search},
author={Reid Pryzant and Dan Iter and Jerry Li and Yin Tat Lee and Chenguang Zhu and Michael Zeng},
booktitle={The 2023 Conference on Empirical Methods in Natural Language Processing},
year={2023}
}

@article{du2024mixture,
	author  = {Du, Hongyang and Liu, Guangyuan and Lin, Yijing and Niyato, Dusit and Kang, Jiawen and Xiong, Zehui and Kim, Dong In},
	journal = {arXiv preprint arXiv:2402.09756},
	title   = {Mixture of Experts for Network Optimization: A Large Language Model-enabled Approach},
	year    = {2024}
}

@article{lee2025convergence,
  title={On the Convergence of Large Language Model Optimizer for Black-Box Network Management},
  author={Lee, Hoon and Zhou, Wentao and Debbah, Merouane and Lee, Inkyu},
  journal={arXiv preprint arXiv:2507.02689},
  year={2025}
}

@article{taboada2025ontology,
  title={Ontology matching with large language models and prioritized depth-first search},
  author={Taboada, Maria and Martinez, Diego and Arideh, Mohammed and Mosquera, Rosa},
  journal={Information Fusion},
  pages={103254},
  year={2025},
  publisher={Elsevier}
}

@inproceedings{jawahar2024llm,
  title={LLM Performance Predictors are good initializers for Architecture Search},
  author={Jawahar, Ganesh and Abdul-Mageed, Muhammad and Lakshmanan, Laks VS and Ding, Dujian},
  booktitle={ACL (Findings)},
  year={2024}
}

@inproceedings{sim2025beyond,
  title={Beyond the hype: Benchmarking llm-evolved heuristics for bin packing},
  author={Sim, Kevin and Renau, Quentin and Hart, Emma},
  booktitle={International Conference on the Applications of Evolutionary Computation (Part of EvoStar)},
  pages={386--402},
  year={2025},
  organization={Springer}
}

@article{huang2024large,
  title={When large language model meets optimization},
  author={Huang, Sen and Yang, Kaixiang and Qi, Sheng and Wang, Rui},
  journal={Swarm and Evolutionary Computation},
  volume={90},
  pages={101663},
  year={2024},
  publisher={Elsevier}
}

@inproceedings{pallagani2024prospects,
  title={On the prospects of incorporating large language models (llms) in automated planning and scheduling (aps)},
  author={Pallagani, Vishal and Muppasani, Bharath Chandra and Roy, Kaushik and Fabiano, Francesco and Loreggia, Andrea and Murugesan, Keerthiram and Srivastava, Biplav and Rossi, Francesca and Horesh, Lior and Sheth, Amit},
  booktitle={Proceedings of the International Conference on Automated Planning and Scheduling},
  volume={34},
  pages={432--444},
  year={2024}
}

@article{du2024llm4ed,
  title={LLM4ED: Large Language Models for Automatic Equation Discovery},
  author={Du, Mengge and Chen, Yuntian and Wang, Zhongzheng and Nie, Longfeng and Zhang, Dongxiao},
  journal={arXiv preprint arXiv:2405.07761},
  year={2024}
}

@article{hu2022exploring,
  title={Exploring evolution-aware \&-free protein language models as protein function predictors},
  author={Hu, Mingyang and Yuan, Fajie and Yang, Kevin and Ju, Fusong and Su, Jin and Wang, Hui and Yang, Fei and Ding, Qiuyang},
  journal={Advances in Neural Information Processing Systems},
  volume={35},
  pages={38873--38884},
  year={2022}
}

@article{buehler2024x,
  title={X-LoRA: Mixture of low-rank adapter experts, a flexible framework for large language models with applications in protein mechanics and molecular design},
  author={Buehler, Eric L and Buehler, Markus J},
  journal={APL Machine Learning},
  volume={2},
  number={2},
  year={2024},
  publisher={AIP Publishing}
}

@article{tran2024protein,
  title={Protein design by directed evolution guided by large language models},
  author={Tran, Thanh VT and Hy, Truong Son},
  journal={IEEE Transactions on Evolutionary Computation},
  year={2024},
  publisher={IEEE}
}

@article{ye2023drugassist,
  title={Drugassist: A large language model for molecule optimization},
  author={Ye, Geyan and Cai, Xibao and Lai, Houtim and Wang, Xing and Huang, Junhong and Wang, Longyue and Liu, Wei and Zeng, Xiangxiang},
  journal={Briefings in Bioinformatics},
  volume={26},
  number={1},
  pages={bbae693},
  year={2025},
  publisher={Oxford University Press}
}

@article{ye2024novo,
  title={De novo drug design as GPT language modeling: large chemistry models with supervised and reinforcement learning},
  author={Ye, Gavin},
  journal={Journal of Computer-Aided Molecular Design},
  volume={38},
  number={1},
  pages={20},
  year={2024},
  publisher={Springer}
}

@article{wang2024catalm,
  title={CataLM: empowering catalyst design through large language models},
  author={Wang, Ludi and Chen, Xueqing and Du, Yi and Zhou, Yuanchun and Gao, Yang and Cui, Wenjuan},
  journal={International Journal of Machine Learning and Cybernetics},
  pages={1--11},
  year={2025},
  publisher={Springer}
}

@inproceedings{rankovic2023bochemian,
  title={BoChemian: Large language model embeddings for Bayesian optimization of chemical reactions},
  author={Rankovi{\'c}, Bojana and Schwaller, Philippe},
  booktitle={NeurIPS 2023 Workshop on Adaptive Experimental Design and Active Learning in the Real World},
  year={2023}
}

@inproceedings{
kristiadi2024sober,
title={A Sober Look at {LLM}s for Material Discovery: Are They Actually Good for Bayesian Optimization Over Molecules?},
author={Agustinus Kristiadi and Felix Strieth-Kalthoff and Marta Skreta and Pascal Poupart and Alan Aspuru-Guzik and Geoff Pleiss},
booktitle={Forty-first International Conference on Machine Learning},
year={2024}
}

@article{bhattacharya2024large,
  title={Large Language Models as Molecular Design Engines},
  author={Bhattacharya, Debjyoti and Cassady, Harrison J and Hickner, Michael A and Reinhart, Wesley F},
  journal={Journal of Chemical Information and Modeling},
  year={2024},
  publisher={ACS Publications}
}

@article{ishida2024large,
  title={Large language models open new way of AI-assisted molecule design for chemists},
  author={Ishida, Shoichi and Sato, Tomohiro and Honma, Teruki and Terayama, Kei},
  journal={Journal of Cheminformatics},
  volume={17},
  number={1},
  pages={36},
  year={2025},
  publisher={Springer}
}

@inproceedings{nasir2023llmatic,
  title={Llmatic: neural architecture search via large language models and quality diversity optimization},
  author={Nasir, Muhammad Umair and Earle, Sam and Togelius, Julian and James, Steven and Cleghorn, Christopher},
  booktitle={proceedings of the Genetic and Evolutionary Computation Conference},
  pages={1110--1118},
  year={2024}
}

@book{glover2006handbook,
	author    = {Glover, Fred W and Kochenberger, Gary A},
	publisher = {Springer Science \& Business Media},
	title     = {Handbook of metaheuristics},
	volume    = {57},
	year      = {2006}
}

@article{girotra2023ideas,
  title={Ideas are dimes a dozen: Large language models for idea generation in innovation},
  author={Girotra, Karan and Meincke, Lennart and Terwiesch, Christian and Ulrich, Karl T},
  journal={Available at SSRN 4526071},
  year={2023}
}

@inproceedings{
si2024can,
title={Can {LLM}s Generate Novel Research Ideas? A Large-Scale Human Study with 100+ {NLP} Researchers},
author={Chenglei Si and Diyi Yang and Tatsunori Hashimoto},
booktitle={The Thirteenth International Conference on Learning Representations},
year={2025}
}

@article{allen2023physics,
  title={Physics of language models: Part 3.1, knowledge storage and extraction},
  author={Allen-Zhu, Zeyuan and Li, Yuanzhi},
  journal={arXiv preprint arXiv:2309.14316},
  year={2023}
}

@inproceedings{liu2025large,
  title={Large language model for multiobjective evolutionary optimization},
  author={Liu, Fei and Lin, Xi and Yao, Shunyu and Wang, Zhenkun and Tong, Xialiang and Yuan, Mingxuan and Zhang, Qingfu},
  booktitle={International Conference on Evolutionary Multi-Criterion Optimization},
  pages={178--191},
  year={2025},
  organization={Springer}
}

@inproceedings{sheng2025solsearch,
  title={SolSearch: An LLM-Driven Framework for Efficient SAT-Solving Code Generation},
  author={Sheng, Junjie and Lin, Yanqiu and Wu, Jiehao and Huang, Yanhong and Shi, Jianqi and Zhang, Min and Wang, Xiangfeng},
  booktitle={2025 IEEE/ACM 47th International Conference on Software Engineering: New Ideas and Emerging Results (ICSE-NIER)},
  pages={6--10},
  year={2025},
  organization={IEEE}
}

@article{yao2025evolution,
  title={Evolution of Optimization Algorithms for Global Placement via Large Language Models},
  author={Yao, Xufeng and Jiang, Jiaxi and Zhao, Yuxuan and Liao, Peiyu and Lin, Yibo and Yu, Bei},
  journal={arXiv preprint arXiv:2504.17801},
  year={2025}
}

@article{ling2025complex,
  title={Complex LLM planning via automated heuristics discovery},
  author={Ling, Hongyi and Parashar, Shubham and Khurana, Sambhav and Olson, Blake and Basu, Anwesha and Sinha, Gaurangi and Tu, Zhengzhong and Caverlee, James and Ji, Shuiwang},
  journal={arXiv preprint arXiv:2502.19295},
  year={2025}
}

@article{bomer2025leveraging,
  title={Leveraging large language models to develop heuristics for emerging optimization problems},
  author={B{\"o}mer, Thomas and Koltermann, Nico and Disselnmeyer, Max and D{\"o}rr, Laura and Meyer, Anne},
  journal={arXiv preprint arXiv:2503.03350},
  year={2025}
}

@article{liu2024llm4ad,
  title={Llm4ad: A platform for algorithm design with large language model},
  author={Liu, Fei and Zhang, Rui and Xie, Zhuoliang and Sun, Rui and Li, Kai and Lin, Xi and Wang, Zhenkun and Lu, Zhichao and Zhang, Qingfu},
  journal={arXiv preprint arXiv:2412.17287},
  year={2024}
}

@article{chen2025heurigym,
  title={HeuriGym: An Agentic Benchmark for LLM-Crafted Heuristics in Combinatorial Optimization},
  author={Chen, Hongzheng and Wang, Yingheng and Cai, Yaohui and Hu, Hins and Li, Jiajie and Huang, Shirley and Deng, Chenhui and Liang, Rongjian and Kong, Shufeng and Ren, Haoxing and others},
  journal={arXiv preprint arXiv:2506.07972},
  year={2025}
}

@article{sun2025co,
  title={Co-bench: Benchmarking language model agents in algorithm search for combinatorial optimization},
  author={Sun, Weiwei and Feng, Shengyu and Li, Shanda and Yang, Yiming},
  journal={arXiv preprint arXiv:2504.04310},
  year={2025}
}

@article{xiao2025survey,
  title={A survey of optimization modeling meets llms: Progress and future directions},
  author={Xiao, Ziyang and Xie, Jingrong and Xu, Lilin and Guan, Shisi and Zhu, Jingyan and Han, Xiongwei and Fu, Xiaojin and Yu, WingYin and Wu, Han and Shi, Wei and others},
  journal={arXiv preprint arXiv:2508.10047},
  year={2025}
}

@inproceedings{zhengmonte,
title={Monte Carlo Tree Search for Comprehensive Exploration in {LLM}-Based Automatic Heuristic Design},
author={Zhi Zheng and Zhuoliang Xie and Zhenkun Wang and Bryan Hooi},
booktitle={Forty-second International Conference on Machine Learning},
year={2025}
}

@article{memduhouglu2024enriching,
  title={Enriching building function classification using Large Language Model embeddings of OpenStreetMap Tags},
  author={Memduho{\u{g}}lu, Abdulkadir and Fulman, Nir and Zipf, Alexander},
  journal={Earth Science Informatics},
  pages={1--16},
  year={2024},
  publisher={Springer}
}

@inproceedings{
shojaee2024llm,
title={{LLM}-{SR}: Scientific Equation Discovery via Programming with Large Language Models},
author={Parshin Shojaee and Kazem Meidani and Shashank Gupta and Amir Barati Farimani and Chandan K. Reddy},
booktitle={The Thirteenth International Conference on Learning Representations},
year={2025}
}

@article{wang2024survey,
  title={A survey on large language model based autonomous agents},
  author={Wang, Lei and Ma, Chen and Feng, Xueyang and Zhang, Zeyu and Yang, Hao and Zhang, Jingsen and Chen, Zhiyuan and Tang, Jiakai and Chen, Xu and Lin, Yankai and others},
  journal={Frontiers of Computer Science},
  volume={18},
  number={6},
  pages={186345},
  year={2024},
  publisher={Springer}
}

@article{zhao2023survey,
	author  = {Zhao, Wayne Xin and Zhou, Kun and Li, Junyi and Tang, Tianyi and Wang, Xiaolei and Hou, Yupeng and Min, Yingqian and Zhang, Beichen and Zhang, Junjie and Dong, Zican and others},
	journal = {arXiv preprint arXiv:2303.18223},
	title   = {A survey of large language models},
	year    = {2023}
}

@article{tang2024learn,
  title={Learn to Optimize-A Brief Overview},
  author={Tang, Ke and Yao, Xin},
  journal={National Science Review},
  pages={nwae132},
  year={2024},
  publisher={Oxford University Press}
}

@article{xu2025autoep,
  title={AutoEP: LLMs-Driven Automation of Hyperparameter Evolution for Metaheuristic Algorithms},
  author={Xu, Zhenxing and Zhang, Yizhe and Bao, Weidong and Wang, Hao and Chen, Ming and Ye, Haoran and Jiang, Wenzheng and Yan, Hui and Wang, Ji},
  journal={arXiv preprint arXiv:2509.23189},
  year={2025}
}

@inproceedings{custode2024investigation,
  title={An investigation on the use of Large Language Models for hyperparameter tuning in Evolutionary Algorithms},
  author={Custode, Leonardo Lucio and Caraffini, Fabio and Yaman, Anil and Iacca, Giovanni},
  booktitle={Proceedings of the Genetic and Evolutionary Computation Conference Companion},
  pages={1838--1845},
  year={2024}
}

@inproceedings{pluhacek2023leveraging,
	author    = {Pluhacek, Michal and Kazikova, Anezka and Kadavy, Tomas and Viktorin, Adam and Senkerik, Roman},
	booktitle = {Proceedings of the Companion Conference on Genetic and Evolutionary Computation},
	pages     = {1812--1820},
	title     = {Leveraging Large Language Models for the Generation of Novel Metaheuristic Optimization Algorithms},
	year      = {2023}
}

@article{ramos2023bayesian,
	author  = {Ramos, Mayk Caldas and Michtavy, Shane S and Porosoff, Marc D and White, Andrew D},
	journal = {arXiv preprint arXiv:2304.05341},
	title   = {Bayesian optimization of catalysts with in-context learning},
	year    = {2023}
}

@inproceedings{narin2024evolutionary,
  title={Evolutionary Reward Design and Optimization with Multimodal Large Language Models},
  author={Narin, Ali},
  booktitle={Proceedings of the 3rd Workshop on Advances in Language and Vision Research (ALVR)},
  pages={202--208},
  year={2024}
}

@inproceedings{huang2024multimodal,
  title={How multimodal integration boost the performance of llm for optimization: Case study on capacitated vehicle routing problems},
  author={Huang, Yuxiao and Zhang, Wenjie and Feng, Liang and Wu, Xingyu and Tan, Kay Chen},
  booktitle={2025 IEEE Symposium for Multidisciplinary Computational Intelligence Incubators (MCII)},
  pages={1--7},
  year={2025},
  organization={IEEE}
}

@article{sartori2024metaheuristics,
  title={Metaheuristics and large language models join forces: Towards an integrated optimization approach},
  author={Sartori, Camilo Chac{\'o}n and Blum, Christian and Bistaffa, Filippo and Corominas, Guillem Rodr{\'\i}guez},
  journal={IEEE Access},
  year={2024},
  publisher={IEEE}
}

@article{wasserkrug2024large,
	author  = {Wasserkrug, Segev and Boussioux, Leonard and Hertog, Dick den and Mirzazadeh, Farzaneh and Birbil, Ilker and Kurtz, Jannis and Maragno, Donato},
	journal = {arXiv preprint arXiv:2402.16269},
	title   = {From Large Language Models and Optimization to Decision Optimization CoPilot: A Research Manifesto},
	year    = {2024}
}

@inproceedings{ahmaditeshnizi2024optimus,
  title={OptiMUS: scalable optimization modeling with (MI) LP solvers and large language models},
  author={AhmadiTeshnizi, Ali and Gao, Wenzhi and Udell, Madeleine},
  booktitle={Proceedings of the 41st International Conference on Machine Learning},
  pages={577--596},
  year={2024}
}

@article{tang2024orlm,
  title={Orlm: A customizable framework in training large models for automated optimization modeling},
  author={Huang, Chenyu and Tang, Zhengyang and Hu, Shixi and Jiang, Ruoqing and Zheng, Xin and Ge, Dongdong and Wang, Benyou and Wang, Zizhuo},
  journal={Operations Research},
  year={2025},
  publisher={INFORMS}
}

@article{guan2023can,
  title={Can large language models design accurate label functions?},
  author={Guan, Naiqing and Chen, Kaiwen and Koudas, Nick},
  journal={arXiv preprint arXiv:2311.00739},
  year={2023}
}

@article{zhou2024autoalign,
	author  = {Rui Zhang and
	           Yixin Su and
	           Bayu Distiawan Trisedya and
	           Xiaoyan Zhao and
	           Min Yang and
	           Hong Cheng and
	           Jianzhong Qi},
	journal = {{IEEE} Trans. Knowl. Data Eng.},
	title   = {AutoAlign: Fully Automatic and Effective Knowledge Graph Alignment
	           Enabled by Large Language Models},
	year    = {2024}
}

@inproceedings{zeng2024tnGPS,
	author    = {Junhua Zeng and
	             Chao Li and
	             Zhun Sun and
	             Qibin Zhao and
	             Guoxu Zhou},
	booktitle = {Forty-first International Conference on Machine Learning, {ICML}},
	title     = {tnGPS: Discovering Unknown Tensor Network Structure Search Algorithms
	             via Large Language Models (LLMs)},
	year      = {2024}
}

@inproceedings{chiquier2024evolving,
  title={Evolving interpretable visual classifiers with large language models},
  author={Chiquier, Mia and Mall, Utkarsh and Vondrick, Carl},
  booktitle={European Conference on Computer Vision},
  pages={183--201},
  year={2024},
  organization={Springer}
}

@article{mao2024identify,
	author     = {Jinzhu Mao and
	              Dongyun Zou and
	              Li Sheng and
	              Siyi Liu and
	              Chen Gao and
	              Yue Wang and
	              Yong Li},
	eprinttype = {arXiv},
	title      = {Identify Critical Nodes in Complex Network with Large Language Models},
	year       = {2024}
}

@article{feng2023knowledge,
  title={Knowledge solver: Teaching llms to search for domain knowledge from knowledge graphs},
  author={Feng, Chao and Zhang, Xinyu and Fei, Zichu},
  journal={arXiv preprint arXiv:2309.03118},
  year={2023}
}

@inproceedings{chen2024hin,
	author    = {Lin Chen and
	             Fengli Xu and
	             Nian Li and
	             Zhenyu Han and
	             Meng Wang and
	             Yong Li and
	             Pan Hui},
	booktitle = {Proceedings of the 30th {ACM} {SIGKDD} Conference on Knowledge Discovery
	             and Data Mining, {KDD}},
	publisher = {{ACM}},
	title     = {Large Language Model-driven Meta-structure Discovery in Heterogeneous
	             Information Network},
	year      = {2024}
}

@article{Ishibashi2024selforganized,
	author     = {Yoichi Ishibashi and
	              Yoshimasa Nishimura},
	eprinttype = {arXiv},
	journal    = {CoRR},
	title      = {Self-Organized Agents: {A} {LLM} Multi-Agent Framework toward Ultra
	              Large-Scale Code Generation and Optimization},
	year       = {2024}
}

@article{liu2025interpretable,
  title={An Interpretable Automated Mechanism Design Framework with Large Language Models},
  author={Liu, Jiayuan and Guo, Mingyu and Conitzer, Vincent},
  journal={arXiv preprint arXiv:2502.12203},
  year={2025}
}

@article{zhang2025llm,
  title={Llm-driven instance-specific heuristic generation and selection},
  author={Zhang, Shaofeng and Liu, Shengcai and Lu, Ning and Wu, Jiahao and Liu, Ji and Ong, Yew-Soon and Tang, Ke},
  journal={arXiv preprint arXiv:2506.00490},
  year={2025}
}

@inproceedings{singh2024enhancing,
  title={Enhancing decision-making in optimization through llm-assisted inference: A neural networks perspective},
  author={Singh, Gaurav and Bali, Kavitesh Kumar},
  booktitle={2024 International Joint Conference on Neural Networks (IJCNN)},
  pages={1--7},
  year={2024},
  organization={IEEE}
}

@inproceedings{wang2024large,
  title={Large Language Model-Aided Evolutionary Search for Constrained Multiobjective Optimization},
  author={Wang, Zeyi and Liu, Songbai and Chen, Jianyong and Tan, Kay Chen},
  booktitle={International Conference on Intelligent Computing},
  pages={218--230},
  year={2024},
  organization={Springer}
}

@inproceedings{aglietti2025funbo,
  title={FunBO: Discovering Acquisition Functions for Bayesian Optimization with FunSearch},
  author={Aglietti, Virginia and Ktena, Ira and Schrouff, Jessica and Sgouritsa, Eleni and Ruiz, Francisco and Malek, Alan and Bellot, Alexis and Chiappa, Silvia},
  booktitle={Forty-second International Conference on Machine Learning},
	year      = {2025}
}

@inproceedings{liu2024largelmea,
  title={Large language models as evolutionary optimizers},
  author={Liu, Shengcai and Chen, Caishun and Qu, Xinghua and Tang, Ke and Ong, Yew-Soon},
  booktitle={2024 IEEE Congress on Evolutionary Computation (CEC)},
  pages={1--8},
  year={2024},
  organization={IEEE}
}

@article{liu2025language,
  title={Language model evolutionary algorithms for recommender systems: Benchmarks and algorithm comparisons},
  author={Liu, Jiao and Sun, Zhu and Feng, Shanshan and Chen, Caishun and Ong, Yew-Soon},
  journal={IEEE Transactions on Evolutionary Computation},
  year={2025},
  publisher={IEEE}
}

@article{pan2024quantum,
  title={Quantum many-body physics calculations with large language models},
  author={Pan, Haining and Mudur, Nayantara and Taranto, William and Tikhanovskaya, Maria and Venugopalan, Subhashini and Bahri, Yasaman and Brenner, Michael P and Kim, Eun-Ah},
  journal={Communications Physics},
  volume={8},
  number={1},
  pages={49},
  year={2025},
  publisher={Nature Publishing Group UK London}
}

@article{du2024large,
  title={Large language models for automatic equation discovery of nonlinear dynamics},
  author={Du, Mengge and Chen, Yuntian and Wang, Zhongzheng and Nie, Longfeng and Zhang, Dongxiao},
  journal={Physics of Fluids},
  volume={36},
  number={9},
  year={2024},
  publisher={AIP Publishing}
}

@article{ni2024mechagents,
  title={MechAgents: Large language model multi-agent collaborations can solve mechanics problems, generate new data, and integrate knowledge},
  author={Ni, Bo and Buehler, Markus J},
  journal={Extreme Mechanics Letters},
  volume={67},
  pages={102131},
  year={2024},
  publisher={Elsevier}
}

@article{ma2024llamoco,
  title={LLaMoCo: Instruction Tuning of Large Language Models for Optimization Code Generation},
  author={Ma, Zeyuan and Guo, Hongshu and Chen, Jiacheng and Peng, Guojun and Cao, Zhiguang and Ma, Yining and Gong, Yue-Jiao},
  journal={arXiv preprint arXiv:2403.01131},
  year={2024}
}

@article{gao2023strategyllm,
  title={Strategyllm: Large language models as strategy generators, executors, optimizers, and evaluators for problem solving},
  author={Gao, Chang and Jiang, Haiyun and Cai, Deng and Shi, Shuming and Lam, Wai},
  journal={Advances in Neural Information Processing Systems},
  volume={37},
  pages={96797--96846},
  year={2024}
}

@article{fan2020difficulty,
  title={Difficulty adjustable and scalable constrained multiobjective test problem toolkit},
  author={Fan, Zhun and Li, Wenji and Cai, Xinye and Li, Hui and Wei, Caimin and Zhang, Qingfu and Deb, Kalyanmoy and Goodman, Erik},
  journal={Evolutionary computation},
  volume={28},
  number={3},
  pages={339--378},
  year={2020},
  publisher={MIT Press One Rogers Street, Cambridge, MA 02142-1209, USA journals-info~…}
}

@inproceedings{hansen2010black,
  title={Black-box optimization benchmarking of NEWUOA compared to BIPOP-CMA-ES: on the BBOB noiseless testbed},
  author={Hansen, Nikolaus and Ros, Raymond},
  booktitle={Proceedings of the 12th annual conference companion on Genetic and evolutionary computation},
  pages={1519--1526},
  year={2010}
}

@article{de2024iohexperimenter,
  title={Iohexperimenter: Benchmarking platform for iterative optimization heuristics},
  author={de Nobel, Jacob and Ye, Furong and Vermetten, Diederick and Wang, Hao and Doerr, Carola and B{\"a}ck, Thomas},
  journal={Evolutionary Computation},
  pages={1--6},
  year={2024},
  publisher={MIT Press One Rogers Street, Cambridge, MA 02142-1209, USA journals-info~…}
}

@article{frazier2016bayesian,
  title={Bayesian optimization for materials design},
  author={Frazier, Peter I and Wang, Jialei},
  journal={Information science for materials discovery and design},
  pages={45--75},
  year={2016},
  publisher={Springer}
}

@inproceedings{zhang2023using,
  title={Using large language models for hyperparameter optimization},
  author={Zhang, Michael R and Desai, Nishkrit and Bae, Juhan and Lorraine, Jonathan and Ba, Jimmy},
  booktitle={NeurIPS 2023 Foundation Models for Decision Making Workshop},
  year={2023}
}

@inproceedings{
  eggensperger2021hpobench,
  title={{HPOB}ench: A Collection of Reproducible Multi-Fidelity Benchmark Problems for {HPO}},
  author={Katharina Eggensperger and Philipp M{\"u}ller and Neeratyoy Mallik and Matthias Feurer and Rene Sass and Aaron Klein and Noor Awad and Marius Lindauer and Frank Hutter},
  booktitle={Thirty-fifth Conference on Neural Information Processing Systems Datasets and Benchmarks Track (Round 2)},
  year={2021}
}

@article{dinh2022lift,
  title={Lift: Language-interfaced fine-tuning for non-language machine learning tasks},
  author={Dinh, Tuan and Zeng, Yuchen and Zhang, Ruisu and Lin, Ziqian and Gira, Michael and Rajput, Shashank and Sohn, Jy-yong and Papailiopoulos, Dimitris and Lee, Kangwook},
  journal={Advances in Neural Information Processing Systems},
  volume={35},
  pages={11763--11784},
  year={2022}
}

@inproceedings{li2021prefix,
  title={Prefix-Tuning: Optimizing Continuous Prompts for Generation},
  author={Li, Xiang Lisa and Liang, Percy},
  booktitle={Proceedings of the 59th Annual Meeting of the Association for Computational Linguistics and the 11th International Joint Conference on Natural Language Processing (Volume 1: Long Papers)},
  pages={4582--4597},
  year={2021}
}

@inproceedings{honovich2022instruction,
  title={Instruction Induction: From Few Examples to Natural Language Task Descriptions},
  author={Honovich, Or and Shaham, Uri and Bowman, Samuel and Levy, Omer},
  booktitle={Proceedings of the 61st Annual Meeting of the Association for Computational Linguistics (Volume 1: Long Papers)},
  pages={1935--1952},
  year={2023}
}

@article{suzgun2022challenging,
  title={Challenging big-bench tasks and whether chain-of-thought can solve them},
  author={Suzgun, Mirac and Scales, Nathan and Sch{\"a}rli, Nathanael and Gehrmann, Sebastian and Tay, Yi and Chung, Hyung Won and Chowdhery, Aakanksha and Le, Quoc V and Chi, Ed H and Zhou, Denny and others},
  journal={arXiv preprint arXiv:2210.09261},
  year={2022}
}

@article{cobbe2021training,
  title={Training verifiers to solve math word problems},
  author={Cobbe, Karl and Kosaraju, Vineet and Bavarian, Mohammad and Chen, Mark and Jun, Heewoo and Kaiser, Lukasz and Plappert, Matthias and Tworek, Jerry and Hilton, Jacob and Nakano, Reiichiro and others},
  journal={arXiv preprint arXiv:2110.14168},
  year={2021}
}

@inproceedings{liu2024largellambo,
title={Large Language Models to Enhance Bayesian Optimization},
author={Tennison Liu and Nicol{\'a}s Astorga and Nabeel Seedat and Mihaela van der Schaar},
booktitle={The Twelfth International Conference on Learning Representations},
year={2024}
}

@inproceedings{ramamonjison2023nl4opt,
  title={Nl4opt competition: Formulating optimization problems based on their natural language descriptions},
  author={Ramamonjison, Rindranirina and Yu, Timothy and Li, Raymond and Li, Haley and Carenini, Giuseppe and Ghaddar, Bissan and He, Shiqi and Mostajabdaveh, Mahdi and Banitalebi-Dehkordi, Amin and Zhou, Zirui and others},
  booktitle={NeurIPS 2022 Competition Track},
  pages={189--203},
  year={2023},
  organization={PMLR}
}

@article{sun2024llm,
  title={LLM-based Multi-Agent Reinforcement Learning: Current and Future Directions},
  author={Sun, Chuanneng and Huang, Songjun and Pompili, Dario},
  journal={arXiv preprint arXiv:2405.11106},
  year={2024}
}

@article{liu2024agentlite,
  title={AgentLite: A Lightweight Library for Building and Advancing Task-Oriented LLM Agent System},
  author={Liu, Zhiwei and Yao, Weiran and Zhang, Jianguo and Yang, Liangwei and Liu, Zuxin and Tan, Juntao and Choubey, Prafulla K and Lan, Tian and Wu, Jason and Wang, Huan and others},
  journal={arXiv preprint arXiv:2402.15538},
  year={2024}
}

@inproceedings{zhou2024design,
  title={Design principle transfer in neural architecture search via large language models},
  author={Zhou, Xun and Wu, Xingyu and Feng, Liang and Lu, Zhichao and Tan, Kay Chen},
  booktitle={Proceedings of the AAAI Conference on Artificial Intelligence},
  volume={39},
  number={21},
  pages={23000--23008},
  year={2025}
}

@inproceedings{yao2024multi,
  title={Multi-objective evolution of heuristic using large language model},
  author={Yao, Shunyu and Liu, Fei and Lin, Xi and Lu, Zhichao and Wang, Zhenkun and Zhang, Qingfu},
  booktitle={Proceedings of the AAAI Conference on Artificial Intelligence},
  volume={39},
  number={25},
  pages={27144--27152},
  year={2025}
}

@inproceedings{wang2024ts,
  title={TS-EoH: An Edge Server Task Scheduling Algorithm Based on Evolution of Heuristic},
  author={Wang, Yatong and Pei, Yuchen and Zhao, Yuqi},
  booktitle={2024 IEEE International Symposium on Parallel and Distributed Processing with Applications (ISPA)},
  pages={693--700},
  year={2024},
  organization={IEEE}
}

@article{liu2025fitness,
  title={Fitness landscape of large language model-assisted automated algorithm search},
  author={Liu, Fei and Zhang, Qingfu and Shi, Jialong and Tong, Xialiang and Mao, Kun and Yuan, Mingxuan},
  journal={arXiv preprint arXiv:2504.19636},
  year={2025}
}

@article{novikov2025alphaevolve,
  title={AlphaEvolve: A coding agent for scientific and algorithmic discovery},
  author={Novikov, Alexander and V{\~u}, Ng{\^a}n and Eisenberger, Marvin and Dupont, Emilien and Huang, Po-Sen and Wagner, Adam Zsolt and Shirobokov, Sergey and Kozlovskii, Borislav and Ruiz, Francisco JR and Mehrabian, Abbas and others},
  journal={arXiv preprint arXiv:2506.13131},
  year={2025}
}

@inproceedings{van2025code,
  title={Code evolution graphs: Understanding large language model driven design of algorithms},
  author={van Stein, Niki and V. Kononova, Anna and Kotthoff, Lars and B{\"a}ck, Thomas},
  booktitle={Proceedings of the Genetic and Evolutionary Computation Conference},
  pages={943--951},
  year={2025}
}

@article{
liu2025eoh,
  title={EoH-S: Evolution of heuristic set using llms for automated heuristic design},
  author={Liu, Fei and Liu, Yilu and Zhang, Qingfu and Tong, Xialiang and Yuan, Mingxuan},
year = {2026},
journal= {Proceedings of the AAAI Conference on Artificial Intelligence},
}

@article{li2025llm,
  title={LLM-Assisted Automatic Memetic Algorithm for Lot-Streaming Hybrid Job Shop Scheduling With Variable Sublots},
  author={Li, Rui and Wang, Ling and Sang, Hongyan and Yao, Lizhong and Pan, Lijun},
  journal={IEEE Transactions on Evolutionary Computation},
  year={2025},
  publisher={IEEE}
}

@inproceedings{wang2025llm,
  title={LLM-enhanced score function evolution for causal structure learning},
  author={Wang, Zidong and Liu, Fei and Feng, Qi and Zhang, Qingfu and Gao, Xiaoguang},
  booktitle={Proceedings of the Thirty-Fourth International Joint Conference on Artificial Intelligence},
  pages={9086--9094},
  year={2025}
}

@inproceedings{xie2025llm,
  title={LLM-Driven Neighborhood Search for Efficient Heuristic Design},
  author={Xie, Zhuoliang and Liu, Fei and Wang, Zhenkun and Zhang, Qingfu},
  booktitle={2025 IEEE Congress on Evolutionary Computation (CEC)},
  pages={1--8},
  year={2025},
  organization={IEEE}
}

@article{shi2025generalizable,
  title={Generalizable heuristic generation through large language models with meta-optimization},
  author={Shi, Yiding and Zhou, Jianan and Song, Wen and Bi, Jieyi and Wu, Yaoxin and Zhang, Jie},
  journal={arXiv preprint arXiv:2505.20881},
  year={2025}
}

@inproceedings{van2025blade,
  title={BLADE: Benchmark suite for LLM-driven Automated Design and Evolution of iterative optimisation heuristics},
  author={van Stein, Niki and V. Kononova, Anna and Yin, Haoran and B{\"a}ck, Thomas},
  booktitle={Proceedings of the Genetic and Evolutionary Computation Conference Companion},
  pages={2336--2344},
  year={2025}
}

@inproceedings{wang2023review,
  title={A review on code generation with llms: Application and evaluation},
  author={Wang, Jianxun and Chen, Yixiang},
  booktitle={2023 IEEE International Conference on Medical Artificial Intelligence (MedAI)},
  pages={284--289},
  year={2023},
  organization={IEEE}
}

@article{
zhao2023automated,
title={Automated Design of Metaheuristic Algorithms: A Survey},
author={Qi Zhao and Qiqi Duan and Bai Yan and Shi Cheng and Yuhui Shi},
journal={Transactions on Machine Learning Research},
issn={2835-8856},
year={2024}
}

@article{zheng2023survey,
  title={A survey of large language models for code: Evolution, benchmarking, and future trends},
  author={Zheng, Zibin and Ning, Kaiwen and Wang, Yanlin and Zhang, Jingwen and Zheng, Dewu and Ye, Mingxi and Chen, Jiachi},
  journal={arXiv preprint arXiv:2311.10372},
  year={2023}
}

@inproceedings{ahn2024large,
  title={Large Language Models for Mathematical Reasoning: Progresses and Challenges},
  author={Ahn, Janice and Verma, Rishu and Lou, Renze and Liu, Di and Zhang, Rui and Yin, Wenpeng},
  booktitle={Proceedings of the 18th Conference of the European Chapter of the Association for Computational Linguistics: Student Research Workshop},
  pages={225--237},
  year={2024}
}

@inproceedings{guo2024large,
  title={Large language model based multi-agents: a survey of progress and challenges},
  author={Guo, Taicheng and Chen, Xiuying and Wang, Yaqi and Chang, Ruidi and Pei, Shichao and Chawla, Nitesh V and Wiest, Olaf and Zhang, Xiangliang},
  booktitle={Proceedings of the Thirty-Third International Joint Conference on Artificial Intelligence},
  pages={8048--8057},
  year={2024}
}

@article{joel2024survey,
  title={A survey on llm-based code generation for low-resource and domain-specific programming languages},
  author={Joel, Sathvik and Wu, Jie and Fard, Fatemeh},
  journal={ACM Transactions on Software Engineering and Methodology},
  year={2024},
  publisher={ACM New York, NY}
}

@article{chen2025hifo,
  title={HiFo-Prompt: Prompting with Hindsight and Foresight for LLM-based Automatic Heuristic Design},
  author={Chen, Chentong and Zhong, Mengyuan and Sun, Jianyong and Fan, Ye and Shi, Jialong},
  journal={arXiv preprint arXiv:2508.13333},
  year={2025}
}

@inproceedings{
hu2025partition,
title={Partition to Evolve: Niching-enhanced Evolution with {LLM}s for Automated Algorithm Discovery},
author={Qinglong Hu and Qingfu Zhang},
booktitle={The Thirty-ninth Annual Conference on Neural Information Processing Systems},
year={2025}
}

@inproceedings{dat2025hsevo,
  title={Hsevo: Elevating automatic heuristic design with diversity-driven harmony search and genetic algorithm using llms},
  author={Dat, Pham Vu Tuan and Doan, Long and Binh, Huynh Thi Thanh},
  booktitle={Proceedings of the AAAI Conference on Artificial Intelligence},
  volume={39},
  number={25},
  pages={26931--26938},
  year={2025}
}

@article{fan2024artificial,
  title={Artificial intelligence for operations research: Revolutionizing the operations research process},
  author={Fan, Zhenan and Ghaddar, Bissan and Wang, Xinglu and Xing, Linzi and Zhang, Yong and Zhou, Zirui},
  journal={arXiv preprint arXiv:2401.03244},
  year={2024}
}

@article{ferrag2025llm,
  title={From llm reasoning to autonomous ai agents: A comprehensive review},
  author={Ferrag, Mohamed Amine and Tihanyi, Norbert and Debbah, Merouane},
  journal={arXiv preprint arXiv:2504.19678},
  year={2025}
}

@article{zhang2025systematic,
  title={A Systematic Survey on Large Language Models for Evolutionary Optimization: From Modeling to Solving},
  author={Zhang, Yisong and Cheng, Ran and Yi, Guoxing and Tan, Kay Chen},
  journal={arXiv preprint arXiv:2509.08269},
  year={2025}
}

@article{ma2025toward,
  title={Toward automated algorithm design: A survey and practical guide to meta-black-box-optimization},
  author={Ma, Zeyuan and Guo, Hongshu and Gong, Yue-Jiao and Zhang, Jun and Tan, Kay Chen},
  journal={IEEE Transactions on Evolutionary Computation},
  year={2025},
  publisher={IEEE}
}

@article{guo2025deepseek,
  title={Deepseek-r1: Incentivizing reasoning capability in llms via reinforcement learning},
  author={Guo, Daya and Yang, Dejian and Zhang, Haowei and Song, Junxiao and Zhang, Ruoyu and Xu, Runxin and Zhu, Qihao and Ma, Shirong and Wang, Peiyi and Bi, Xiao and others},
  journal={arXiv preprint arXiv:2501.12948},
  year={2025}
}

@article{da2025large,
  title={Large language models for combinatorial optimization: A systematic review},
  author={Da Ros, Francesca and Soprano, Michael and Di Gaspero, Luca and Roitero, Kevin},
  journal={arXiv preprint arXiv:2507.03637},
  year={2025}
}

\end{document}